\documentclass[sn-nature]{sn-jnl}

\usepackage{graphicx}%
\usepackage{multirow}%
\usepackage{amsmath,amssymb,amsfonts}%
\usepackage{amsthm}%
\usepackage{mathrsfs}%
\usepackage[title]{appendix}%
\usepackage{xcolor}%
\usepackage{textcomp}%
\usepackage{manyfoot}%
\usepackage{booktabs}%
\usepackage{algorithm}%
\usepackage{algorithmicx}%
\usepackage{algpseudocode}%
\usepackage{listings}%
\usepackage{subcaption}
\usepackage{tabularx}
\usepackage{makecell}
\usepackage{lscape}

\makeatletter

\makeatletter

\theoremstyle{thmstyleone}%
\theoremstyle{thmstyletwo}%

\theoremstyle{thmstylethree}%

\begin{document}

\title[Article Title]{Local neural operator for solving transient partial differential equations on varied domains}


\author*[1]{\fnm{Hongyu} \sur{Li}}\email{973wtj@xjtu.edu.cn}
\equalcont{These authors contributed equally to this work.}

\author[2]{\fnm{Ximeng} \sur{Ye}}\email{yeximeng@stu.xjtu.edu.cn}
\equalcont{These authors contributed equally to this work.}

\author[1]{\fnm{Peng} \sur{Jiang}}\email{jiangpeng219@xjtu.edu.cn}
\author*[2]{\fnm{Guoliang} \sur{Qin}}\email{glqin@xjtu.edu.cn}
\author*[1]{\fnm{Tiejun} \sur{Wang}}\email{lihongyu@stu.xjtu.edu.cn}

\affil[1]{\orgdiv{State Key Lab for Strength and Vibration of Mechanical Structures, Department of Engineering Mechanics}, \orgname{Xi’an Jiaotong University}, \orgaddress{\city{Xi'an}, \postcode{710049}, \state{Shaanxi}, \country{China}}}

\affil[2]{\orgdiv{School of Energy and Power Engineering}, \orgname{Xi’an Jiaotong University}, \orgaddress{\city{Xi'an}, \postcode{710049}, \state{Shaanxi}, \country{China}}}


\abstract{Artificial intelligence (AI) shows great potential to reduce the huge cost of solving partial differential equations (PDEs).
However, it is not fully realized in practice as neural networks are defined and trained on fixed domains and boundaries. 
Herein, we propose local neural operator (LNO) for solving transient PDEs on varied domains.
It comes together with a handy strategy including boundary treatments, enabling one pre-trained LNO to predict solutions on different domains.
For demonstration, LNO learns Navier-Stokes equations from randomly generated data samples, and then the pre-trained LNO is used as an explicit numerical time-marching scheme to solve the flow of fluid on unseen domains, e.g., the flow in a lid-driven cavity and the flow across the cascade of airfoils.
It is about 1000$\times$ faster than the conventional finite element method to calculate the flow across the cascade of airfoils. 
The solving process with pre-trained LNO achieves great efficiency, with significant potential to accelerate numerical calculations in practice.}




\maketitle

\section{Introduction}\label{sec1}

The physical laws of fluid flow, heat transfer, wave propagation, etc., are important for human health, sports, environment management, modern industry and engineering.
So, various transient partial differential equations (PDEs) are formulated to describe the physical laws.
However, it is not easy to solve them in practice, especially for the non-linear ones such as Navier-Stokes (N-S) equations, etc. 
Although many numerical schemes have been proposed and hardware such as supercomputers are built up for large-scale computations, the huge costs in time and money could be unaffordable in some of the scientific and/or engineering practices. 
The latest artificial intelligence (AI) shows great potential to accelerate the solving process of PDEs \cite{Bar-Sinai2019,Dmitrii2021}. The process can be accelerated hundreds of times faster by using neural networks as direct solvers substituting the conventional ones \cite{LiZongyi2021}.
Unfortunately, the current  AI substitutes do not fully meet the applications because one has to train neural networks for each computational domain.
This limitation is due to the fixed computational domain assumed in the learning problem definition.
Hence, we have to rethink ‘what to learn’ for neural networks regarding the issue of reusability in different computational domains.

The journey of developing AI method for solving PDEs is substantially the process of exploring and discussing ‘what to learn’, i.e., the approximating target for neural networks.
Early attempts are to use neural networks to approximate the solution function $u(x,t)$ by minimizing the residual of PDEs \cite{Lagaris1998,Psichogios1992}, i.e., the input is vectors representing the positions and the output is values at these points.
Physics-informed neural networks (PINNs) \cite{Raissi2019,Raissi2020}, deep Galerkin method \cite{Sirignano2018}, and deep Ritz method \cite{Weinan2018} developed this idea and earned attention \cite{Chen2021,Wang2022,Wang2020}.
More recently, there are promising approaches to approximate operators that, the input and output of the neural network are conceptually generalized as vectors with infinite dimensions, i.e., the functions.
These models are called neural operators \cite{Kovachki2021}, and the pre-trained neural operator can predict solution functions of PDE in more than one case.
Impressive examples include Deep Operator Network (DeepONet) \cite{LuLu2021}, Fourier Neural Operator (FNO) \cite{LiZongyi2020}, and its several variants \cite{LiZongyi2021,Li2021,Gupta2021} with applications \cite{Jiang2021,Pathak2022}.
However, the varied computational domain troubles the application of these methods. 
Although there is valuable progress on this issue by using novel architecture \cite{Kashefi2022}, transfer learning techniques \cite{Goswami2022}, sophisticated composite algorithms \cite{WangH2022}, and fine-tuning for extrapolation \cite{zhu2023}, it is still an open problem. 

\begin{figure}[htbp]
\centering
\includegraphics[width=1.2\textwidth]{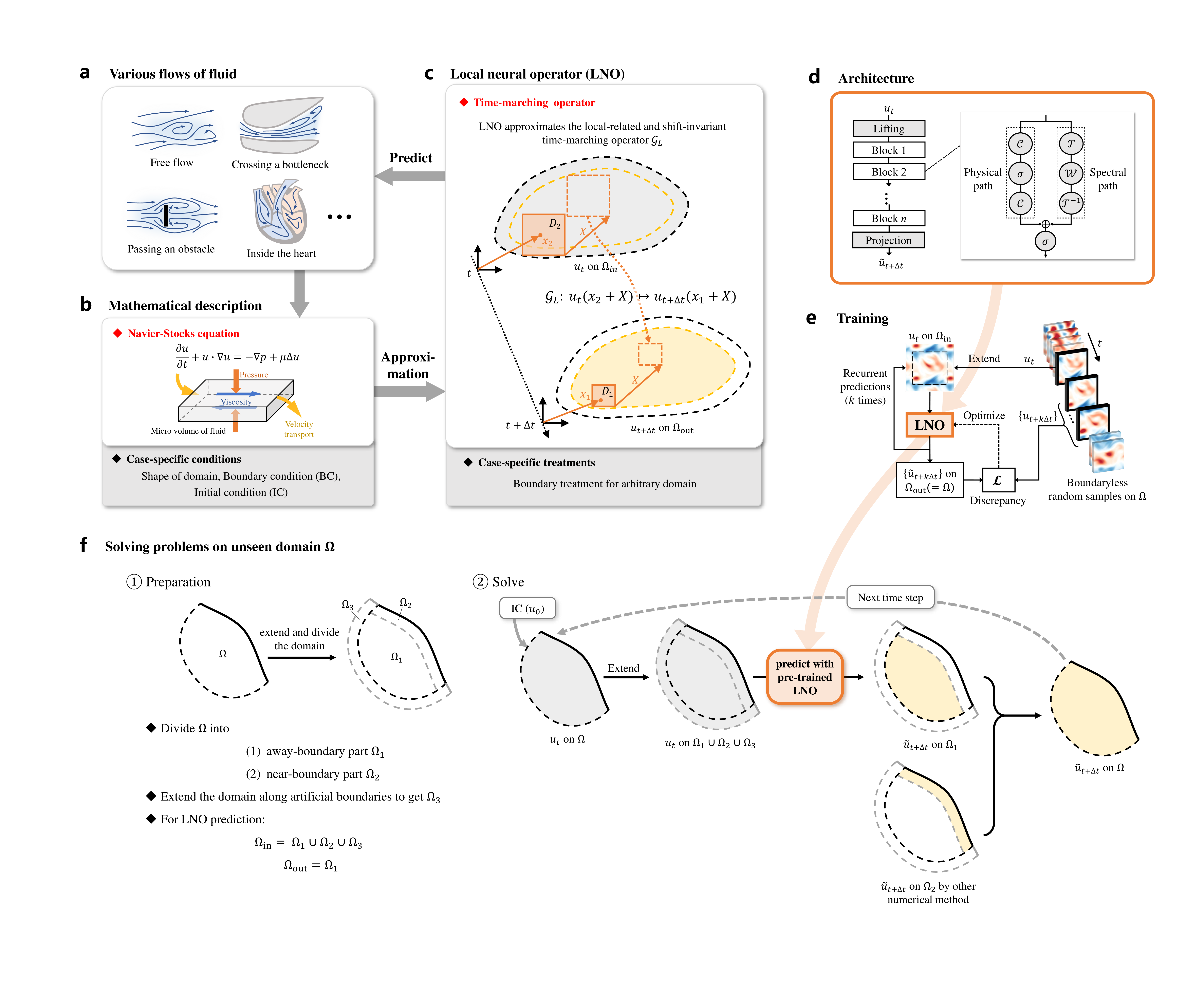}
\setlength{\leftskip}{-30pt}
\caption{\textbf{Local neural operator (LNO) conception and method.}
\textbf{a}, Various flows of fluids.
\textbf{b}, Mathematical description of the flow in different case-specific conditions.
\textbf{c}, The concept of LNO to approximate local-related and shift-invariant time-marching operator $\mathcal{G}_L$ representing transient PDEs, while the case-specific conditions are imposed by specific boundary treatment. 
\textbf{d}, The specific architecture of the LNO, in which $\mathcal{C},\sigma,\mathcal{W}$ are convolutional layers, activation functions, and the linear layer, and $\mathcal{T}$  and $\mathcal{T}^{-1}$ are Legendre spectral transform and its inverse on local parts of the computational domain, respectively.
\textbf{e}, Training LNO with samples on periodic domains while the boundary effect is excluded. After that, the pre-trained LNO can predict solutions on different domains by collaborating with case-specific boundary treatment.
\textbf{f}, Procedure to solve transient PDEs on unseen domains with pre-trained LNO.
The dotted lines are the artificial boundary of the domain, while the solid lines are ordinary ones.
The areas shaded with gray or yellow in c and f are respectively the support set of the input $u_t$ and the output $u_{t+\Delta t}$ .
}\label{fig:LNOconcept}
\end{figure}

This work raises a new learning problem to let neural networks learn transient PDEs separated from case-specific conditions such as the shape of the domain, boundary condition (BC), and initial condition (IC).
The raised problem stands on the fact that, for example, the same fluid performs distinctive flowing patterns while it is in different domains with different boundaries (Fig. \ref{fig:LNOconcept}a).
In view of mathematics, these cases can be described by  identical N-S equations and varied case-specific conditions such as IC, BC, and shape of the computational domain (Fig. \ref{fig:LNOconcept}b).
To learn the equations separately, we propose a \emph{local neural operator} (LNO) to approximate local-related and shift-invariant time-marching operator for transient PDEs (Fig. \ref{fig:LNOconcept}c).
Thus, one pre-trained LNO can solve problems defined on different domains by collaborating with case-specific boundary treatment.

\section{Local Neural Operator (LNO)}\label{sec2}

Time marching is a common way to solve transient PDEs, by which the physical fields are recurrently solved to the next time level.
This solving process can be modeled as a time-marching operator for neural operator learning \cite{LiZongyi2020,Li2021,Lu2022}
\begin{equation}
\mathcal{G}:u_t(x) \mapsto u_{t+\Delta t}(x), \qquad t\geq 0,x\in \Omega,
\label{eq:1}
\end{equation}
which means that the operator $\mathcal{G}$ maps $u_t$ (the physical fields at time $t$) to  $u_{t+\Delta t}$ (the physical fields at time $t+\Delta t$). 
Mathematically, $u_t$ and $u_{t+\Delta t}$  are functions taking values in $\mathbb{R}^{d_u}$.
$d_u$ is the number of physical fields and $\Delta t$ is the time interval. 
$\Omega \subset \mathbb{R}^d$ is the computational domain, and $d$ is the number of dimensions. 

The assumed certain computational domain $\Omega$ in Eq. (\ref{eq:1}) troubles the application of pre-trained neural operators to varied domains.
To make it flexible, we introduce the following two assumptions.
One is \emph{local-related condition}. 
It is natural in classical physics that, within a limited time interval $\Delta t$, $u_t$ at $x_b$ only impacts $u_{t+\Delta t}$ at $x_a$ with limited distance to $x_b$. This leads to the local-related condition,
\begin{equation}
\frac{\partial u_{t+\Delta t}(x_a)}{\partial u_{t}(x_b)}=0,\qquad\forall \left \| x_a-x_b \right \|>r,
\label{eq:2}
\end{equation}
where $r$ is the upper bound of the distance between related $x_a$ and $x_b$. 
The minimum  of $r$, denoted as $r_\textup{min}$, describes the local-related range, which is relatively small if $\Delta t$ is small. 
The other assumption is \emph{shift-invariant condition}.
The time-marching operator stays objective wherever the coordinate is. 
For example, the origin point of the coordinate may move to $X \in \mathbb{R}^d$, and the time-marching operator still stands as the variable $x$ is substituted to $x-X$.

Based on the above assumptions, the time-marching operator can be written as 
\begin{equation}
\mathcal{G}_L:u_t(x_2+X) \mapsto u_{t+\Delta t}(x_1+X),\qquad t\geq 0,x_1 \in D_1,x_2\in D_2,X\in\mathbb{R}^d,
\label{eq:3}
\end{equation}
where $D_2$ and $D_1$ are the unit domains of the input and output functions, respectively. 
 $D_2$ is determined by $D_1$ to ensure $\left \| x_1-x_2 \right \|\leq r_\textup{min}$  according to Eq. (\ref{eq:2}).
 $X$ is the shifting vector. 
With the bounded domain $\chi \subset \mathbb{R}^d$ given according to the computational domain of a certain case, Eq. (\ref{eq:3}) is equivalently

\begin{gather}
    \mathcal{G}_L: u_t\left(x_2^{\prime}\right) \mapsto u_{t+\Delta t}\left(x_1^{\prime}\right), \qquad t \geq 0, x_1^{\prime} \in \Omega_\textup{out}, x_2^{\prime} \in \Omega_\textup{in}, \\
\Omega_\textup{out}=\left\{x_1+X \mid x_1 \in D_1, X \in \chi\right\}, \nonumber\\
\Omega_\textup{in}=\left\{x_2+X \mid x_2 \in D_2, X \in \chi\right\}.\nonumber
\label{eq:4}
\end{gather}
As $\chi$ is variable for different cases, the domain $\Omega_\textup{out}$ and $\Omega_\textup{in}$ are variable.
This means that $\mathcal{G}_L$ is the mapping between functions on varied output domain $\Omega_\textup{out}$ and functions on its corresponding input domain $\Omega_\textup{in}$ (Fig. \ref{fig:LNOconcept}c).

We propose a \emph{local neural operator} (LNO) to approximate the local-related and shift-invariant time-marching operator $\mathcal{G}_L$ defined in Eq.(\ref{eq:4}).
The architecture of LNO (Fig. \ref{fig:LNOconcept}d) follows a common lifting-projection structure \cite{Kovachki2021,LiZongyi2020}.
All specific layers in the present LNO are distinctively designed as local-related.
Specifically, the inner block comprises a physical path and a spectral path to enrich the approximating ability of LNO.
In the physical path, local-related convolutional layers link function values at different positions directly in the physical space.
In the spectral path, the interior functions are transformed in the spectral space with Legendre polynomials as the basis. 
The spectral transform is conducted on a sliding unit window to ensure the output functions are local-related to the input. 
More details about the architecture of LNO is presented in Section \ref{sec5}.

We train the LNO with randomly generated samples following a supervised training scheme (Fig. \ref{fig:LNOconcept}e). 
According to the definition in Eq. (\ref{eq:4}), the samples can be data series $\left \{ u_{k\Delta t}|k\in\mathbb{N} \right \}$ on varied domains.
Here, we specifically generate samples on a square domain $\Omega$ with periodic boundaries.
Before sending $u_t$ into LNO, the input domain is extended to $\Omega_\textup{in}$ to keep the output domain $\Omega_\textup{out}$ identical to the original input domain $\Omega$.
Besides, the LNO is trained as a recurrent neural network (RNN) to make it stable during recurrent time marching. 
In each training iteration, the initial input $u_t$ is randomly sampled from the dataset, and the output $\tilde{u}_{t+\Delta t}$ is recurrently served as the next input to obtain an output series $\left \{ \tilde{u}_{t+k\Delta t} \right \}_{k=1}^{10}$.
Then, the LNO is trained by minimizing the discrepancy between $\left \{ \tilde{u}_{t+k\Delta t} \right \}_{k=1}^{10}$ and the real solutions  $\left \{u_{t+k\Delta t} \right \}_{k=1}^{10}$.
More implementing details are in Section \ref{sec5}.

According to Eq. (\ref{eq:4}) and Fig. \ref{fig:LNOconcept}c, the pre-trained LNO maps $u_t$ on $\Omega_\textup{in}$ to $u_{t+\Delta t}$ on the smaller domain $\Omega_\textup{out}$.
Thus, a proper boundary treatment is required to hold the computational domain unchanged during the time-marching process.
To achieve a general treatment, the boundaries are classified into two types:
1) Boundaries allowing extension, also known as artificial boundary conditions, e.g., the far-field or periodic boundary. 
2) Boundaries that cannot be extended, e.g., the solid wall boundary.
To apply the pre-trained LNO to solve problems on unseen domains, the workflow to march $u_t$ to $\tilde{u}_{t+\Delta t}$ (the approximation of $u_{t+\Delta t}$) on $\Omega$ is shown in Fig. \ref{fig:LNOconcept}f.
Firstly,  $\Omega$ is extended and divided into $\Omega_1$, $\Omega_2$ and $\Omega_3$, where $\Omega=\Omega_1 \cup\Omega_2$ and  $\Omega_3$ is the extension of $\Omega$. 
Then, the pre-trained LNO takes $u_t$ as input on $\Omega_1 \cup\Omega_2\cup\Omega_3$ and $\tilde{u}_{t+\Delta t}$ as output on $\Omega_1$.
$\tilde{u}_{t+\Delta t}$ on $\Omega_2$ is obtained by other numerical methods \cite{Peskin2002,Uhlmann2005}.
Combining $\tilde{u}_{t+\Delta t}$ on $\Omega_1$ and $\Omega_2$, we obtain $\tilde{u}_{t+\Delta t}$  on $\Omega$.
Thus, we complete one time-marching step forward.
For long-term prediction, the pre-trained LNO takes the initial condition $u_0$ as the first input and recurrently predicts the solution $\left \{ u_{k\Delta t}|k\in \mathbb{N} \right \}$.

\section{Results}\label{sec3}

The proposed conception of LNO is demonstrated by solving N-S equations.
Here, we consider the 2-D case of viscous incompressible fluid flow with no external force as
\begin{equation}
\begin{split}
\frac{\partial u(x, t)}{\partial t}+u(x, t) \cdot \nabla u(x, t)&=-\nabla p(x, t)+\mu \Delta u(x, t), \qquad t>0, \\
\nabla \cdot u(x, t)&=0, \qquad t>0, \\
u(x, 0)&=u_0(x),
\label{eq:5}
\end{split}
\end{equation}
where $u\in \mathbb{R}^2$ is the vector field of velocity, $u_0$ is the initial field of $u$, $\mu$ is the viscosity. 
For this task, the LNO predicts $u_{t+\Delta t}$ using $u_t$ as input, regarding the pressure $p$ as an implicit variable.
Reference solutions for examining the LNO predictions are obtained by using finite element method (FEM) numerical calculations (see Appendix \ref{secA1} for details).

\vspace{8mm}

\begin{figure}[htbp]%
\centering
\includegraphics[width=\textwidth]{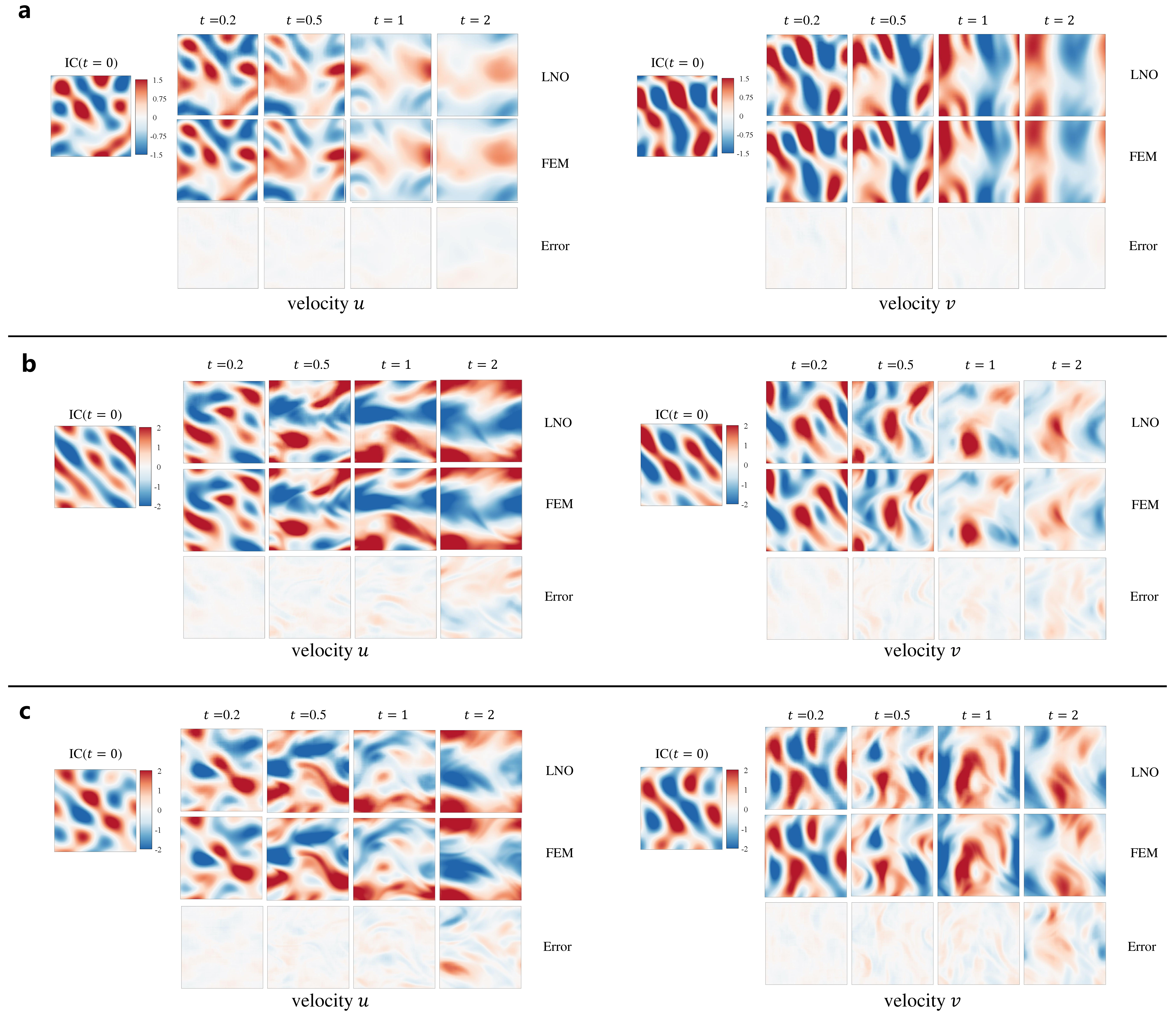}
\caption{Predicted velocity contours by trained LNO for 2-D incompressible N-S equations with three viscosities. \textbf{a},  $\mu=0.01$, \textbf{b},  $\mu=0.002$, \textbf{c},  $\mu=0.001$. 
The trained LNO predicts the solution function for each viscosity according to a random IC different from training data. LNO takes 5, 10, 20, and 40 cycles to predict these four frames at $t=0.2,0.5,1,2$. 
The FEM results are also presented for comparison.}\label{fig:validation}
\end{figure}

\textbf{Validation of LNO training.}
The trained LNO is validated by predicting the free flows generated by 10 ICs which differ from  training sample. 
For velocity fields discretized as $128\times 128$ matrices, the validation accuracy of LNO is described by the mean $L_2$ error of velocity at time $t$,
\begin{equation}
E_t=\frac{1}{10} \sum_{i=1}^{10} \frac{1}{128^2} \sum_{a=1}^{128} \sum_{b=1}^{128}\left\|u_{t, ab}^{(i)}-\tilde{u}_{t, ab}^{(i)}\right\|_2, \quad t=k \Delta t, k \in \mathbb{N}^{+} .
\label{eq:6}
\end{equation}
where $\tilde{u}_t$ is the prediction of LNO, and $u_t$ is the reference solution (the ground truth).
The superscript ‘$(i)$’ denotes the $i^{\textup{th}}$  piece of sample for validation, and the subscript ‘$ab$’ denotes the position of the discretized velocity field.
Three tasks for learning N-S equations with different viscosities ($\mu=0.01,0.002,0.001$) are considered.
Table \ref{tab:1} lists the primary parameters, the number of trainable weights, and $E_t$ at four moments ($t=0.2,0.5,1,2$) of LNOs compared to FNO \cite{LiZongyi2020}.
It is clear that the present LNO owns fewer trainable weights and gets lower error. 
We attribute this improvement to the definition of local-related learning problem, i.e., the finite related range shown in Table \ref{tab:1} ($r_{\textup{min}}=31\Delta x,41\Delta x,61\Delta x$). 
It provides LNO with helpful prior knowledge as some redundant input information is excluded. 
For presenting the results intuitively, the contours of the velocity fields predicted by the present LNO  are shown in Fig. \ref{fig:validation} compared with FEM results. 
It is seen that the results agree well.
Physically, the smaller $\mu$ in N-S equations indicates less viscosity and dissipation, which leads to more complex flowing patterns.
Therefore, it is more difficult for neural networks to learn. 
Still, the present LNO successfully predicts the delicate flowing patterns of small viscosity. 

In what follows, using the trained LNO, we predict the internal flow in a lid-driven cavity and the external flow across the cascade of airfoils to show the reusability of the LNO.

\begin{sidewaystable}
\renewcommand\arraystretch{2}
\begin{center}
\begin{minipage}{\textheight}
\caption{Comparison of the mean $L_2$ error between LNO and FNO in solving 2-D incompressible N-S equations.
The averaged error is shown together with the standard deviation of 10 runs.\label{tab:1}}
\begin{tabular*}{\textheight}{@{\extracolsep{\fill}}cccccccc@{\extracolsep{\fill}}}

\toprule
\multirow{2}{*}{Viscosity}                      & \multirow{2}{*}{Network} & \multirow{2}{*}{\tnote{*}Parameters}                & \multirow{2}{*}{\makecell{\tnote{**}Number of\\trainable\\weights}} & \multicolumn{4}{c}{$E_t$ (mean $L_2$ error at time $t$)}     \\\cmidrule{5-8}
                                          &                        &                                            &                                                & 0.2s   & 0.5s     & 1s     & 2s     \\
\midrule
\multirow{2}{*}{0.01} & FNO \cite{LiZongyi2020}            & $r_{\text{min}}=\infty$                                         &  926326   & 
                      0.062$\pm$0.003  & 0.101$\pm$0.007  & 0.164$\pm$0.021  & 0.209$\pm$0.036 \\\cmidrule{2-8}
                      & The present LNO     &  \makecell{$N=12,M=6,$\\$k=2,r_{\text{min}}=31\Delta x$ }       &  328656   & 0.067$\pm$0.003  & 0.081$\pm$0.006  & 0.132$\pm$0.015  & 0.204$\pm$0.031 \\
\midrule
\multirow{2}{*}{0.002}& FNO \cite{LiZongyi2020}           & $r_{\text{min}}=\infty$                                         & 926326     & 
                      0.097$\pm$0.002  & 0.237$\pm$0.010  & 0.503$\pm$0.032  & 1.013$\pm$0.084 \\\cmidrule{2-8}
                      & The present LNO    &   \makecell{$N=16,M=8,$\\$k=2,r_{\text{min}}=41\Delta x$}       & 776656     & 0.084$\pm$0.004  & 0.166$\pm$0.015  & 0.361$\pm$0.045  & 0.833$\pm$0.089 \\
\midrule
\multirow{2}{*}{0.001}& FNO \cite{LiZongyi2020}           & $r_{\text{min}}=\infty$                                         & 926326     & 
                      0.109$\pm$0.006  & 0.270$\pm$0.015  & 0.603$\pm$0.032  & 1.480$\pm$0.185 \\\cmidrule{2-8}
                      & The present LNO    &  \makecell{$N=24,M=8,$\\$k=2,r_{\text{min}}=61\Delta x$ }       & 776656     & 0.105$\pm$0.007  & 0.241$\pm$0.023  & 0.561$\pm$0.063  & 1.329$\pm$0.127 \\
\bottomrule
\end{tabular*}
\begin{tablenotes}
\item * $N,M,k$ are the window size, the number of adopted modes, and the number of repetitions, respectively.
$r_{\text{min}}=\frac{N}{k}\Delta x+R(n,N,k)$ is the local-related range. $\Delta x=1/64$. See Section \ref{sec5} for $R(n,N,k)$ and more details.
\item ** The complex weights of FNO are counted twice.
\end{tablenotes}
\end{minipage}
\end{center}
\end{sidewaystable}

\vspace{8mm}

\begin{figure}[htbp]%
\centering
\includegraphics[width=0.9\textwidth]{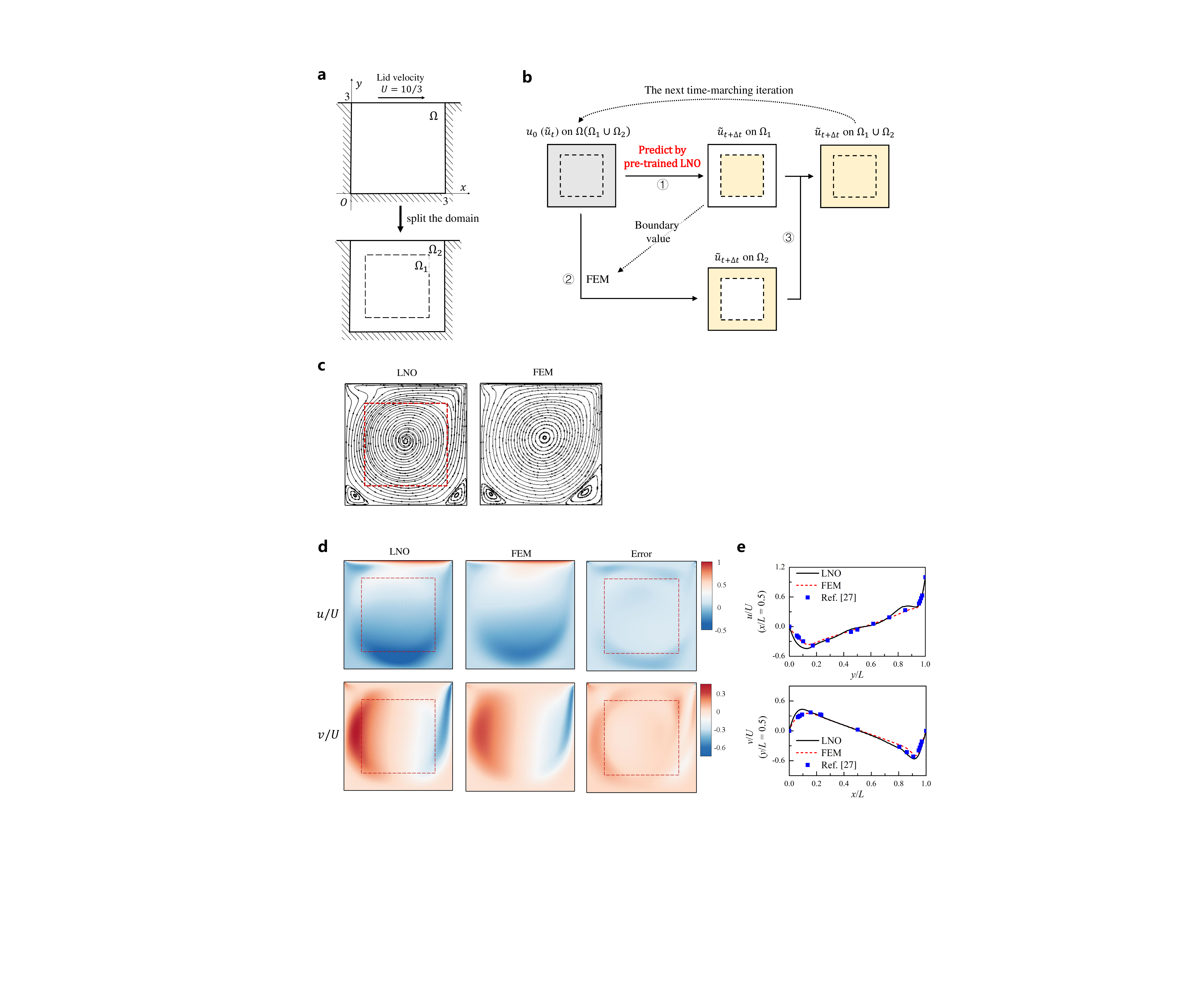}
\caption{\textbf{The pre-trained LNO solves the internal flow in a lid-driven cavity ($Re=1000$).} 
\textbf{a}, Schematics of the problem and the domain division for LNO prediction. 
\textbf{b}, The time-marching workflow to predict the velocity fields with pre-trained LNO.
\textbf{c}, Comparison of the LNO predicted streamlines with FEM numerical results.
\textbf{d} and \textbf{e}, LNO predicted velocity contours and velocity profiles on the centerlines $x/L=0.5$ (the upper) and $y/L=0.5$ (the lower), respectively, in which the results from FEM and Ghia et al. \cite{Ghia1982} are also presented for comparison. }\label{fig:liddriven}
\end{figure}

\textbf{The pre-trained LNO solves the internal flow in a lid-driven cavity.}
The flow in lid-driven cavity is a model problem usually used to test numerical schemes \cite{Ghia1982}, as shown in Fig. \ref{fig:liddriven}a. 
The fluid obeys Eq. (\ref{eq:5}) with viscosity $\mu=0.01$ which is already learned by the pre-trained LNO.
The flow is in a square cavity $\Omega=[0,L]\times[0,L]$ with $L=3$, and is driven by a constant velocity $U=10/3$ on the upper side. 
The other three sides are solid walls with no-slip boundary condition.
The Reynold number $Re=\rho UL/\mu=1000$ (the density $\rho=1$) for this problem.
The cavity $\Omega$ is divided into the away boundary domain $\Omega_1$ (54.7\% area of $\Omega$) and the near-boundary domain $\Omega_2$ according to the corrosion width $R(n,N,k)$ of LNO (see  Section \ref{sec5} for detail).
There is no artificial boundary, so the domain extension is unnecessary. 
The workflow for time marching is shown in Fig. \ref{fig:liddriven}b.
Firstly, LNO takes $u_t$ as the input on $\Omega_1\cup\Omega_2$ and $\tilde{u}_{t+\Delta t}$ as the output on $\Omega_1$.
Secondly,  taking $\tilde{u}_{t+\Delta t}$ on the interface between $\Omega_1$ and $\Omega_2$ predicted by LNO, $\tilde{u}_{t+\Delta t}$ on $\Omega_2$ is calculated by using $Q_2-P_1$ FEM \cite{Brezzi1991}. 
Thirdly, combining $\tilde{u}_{t+\Delta t}$ on $\Omega_1$ and $\Omega_2$, we obtain $\tilde{u}_{t+\Delta t}$ on the complete domain $\Omega$.

LNO takes the IC ($u=0,v=0$) as the first input and predicts $\{\tilde{u}_{k\Delta t}|k\in\mathbb{N}\}$ recurrently until the shown convergent state is reached. 
We compare the steady-state velocity fields predicted by LNO and that solved  by FEM.
Fig. \ref{fig:liddriven}c shows the streamlines, and Fig. \ref{fig:liddriven}d presents the contours of velocities on the domain.
The velocity profiles on centerlines $x/L=0.5$ and $y/L=0.5$ are shown in Fig. \ref{fig:liddriven}e for comparisons of the results of LNO, FEM, and  literature \cite{Ghia1982}. 
It is seen that LNO captures the vortex structure correctly.
A main vortex occupies the center of the cavity and two small vortices are located at the lower left and right corners.
Compared to the FEM numerical solutions, the LNO results show relatively small mean absolute errors as 0.0343 and 0.0302 for the normalized velocities $u/U$ and $v/V$, respectively.
Taking $U$ as a reference, the error rate is lower than 3.5\%. 
In view of efficiency, the implicit FEM costs 9.849 seconds for one step forward on the complete domain $\Omega$ (i.e., $\Omega_1\cup\Omega_2$).
Comparatively, LNO costs only 0.005 seconds on $\Omega_1$, FEM costs 4.575 seconds on $\Omega_2$, and the total time is 4.580 seconds for one step forward on the complete domain $\Omega$.
It is seen that the present LNO has great potential for speeding up numerical analysis. 

\vspace{8mm}

\begin{figure}[htbp]%
\centering
\includegraphics[width=1.3\textwidth]{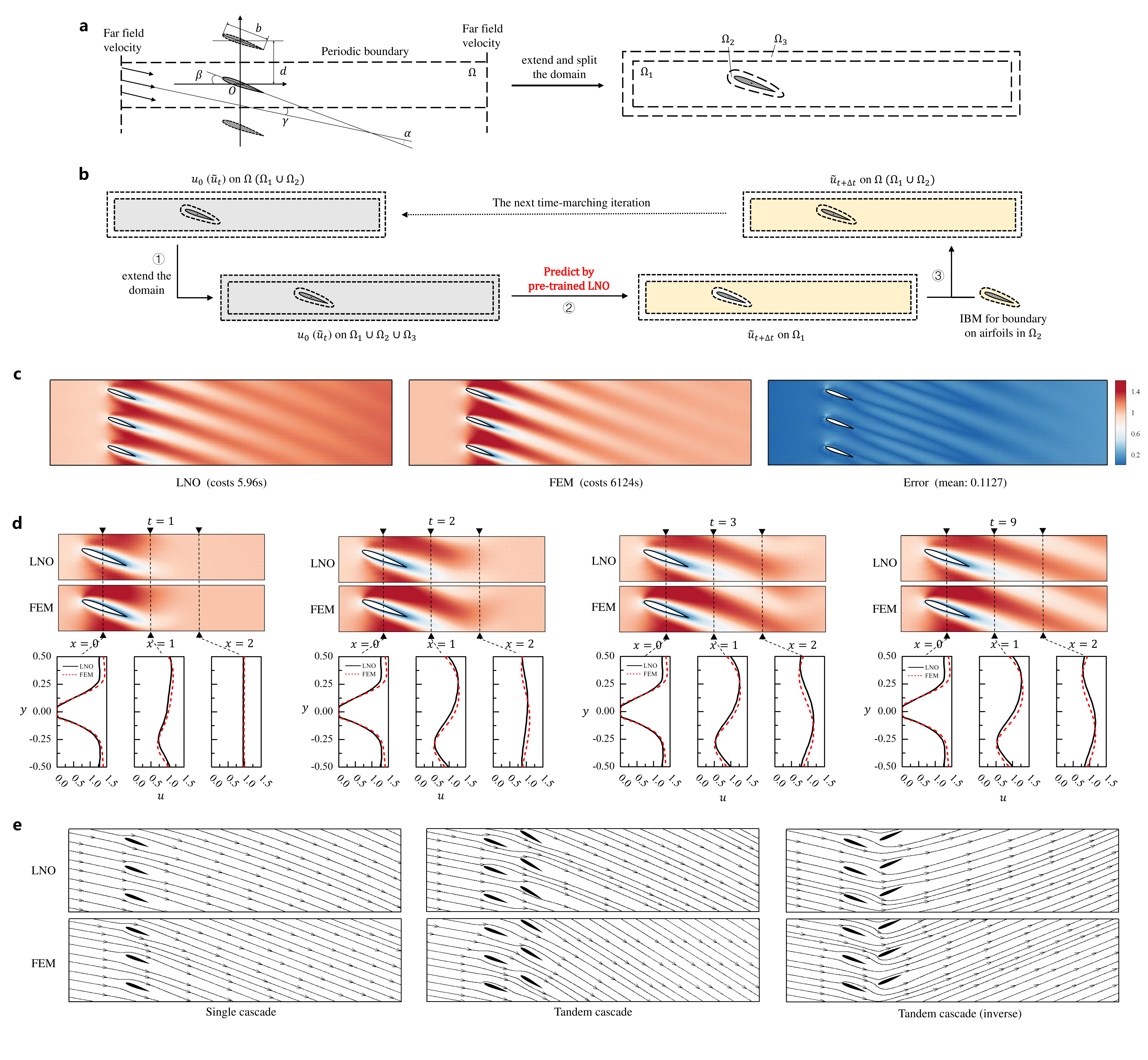}
\setlength{\leftskip}{-60pt}
\caption{\textbf{The pre-trained LNO solves the external flow across the cascade of airfoils.}
\textbf{a}, Schematics of the problem with $b,d,\alpha,\beta,\gamma$ being the chord length, the interval, the angle of attack, the stagger angle, and the inflow angle, respectively. 
The domain is extended and divided to $\Omega_1$, $\Omega_2$ and $\Omega_3$ for LNO prediction.
\textbf{b}, The time-marching workflow to predict the velocity fields with pre-trained LNO.
\textbf{c}, Contours of velocity magnitude and absolute error at the steady state. 
The results are computed on the domain including one airfoil, and are periodically extended to three for better visualization. \textbf{d}, Evolving history of contours of velocity magnitude with velocity profiles at $x=0,1,2$. 
\textbf{e}, Steady-state streamlines predicted by the same pre-trained LNO for flow across different cascades. The FEM results are presented for comparison. }\label{fig:airfoil}
\end{figure}

\textbf{The pre-trained LNO solves the external flow across the cascade of airfoils.}
The external flow around objects is a common problem in engineering \cite{Sekar2019,Bhatnagar2019}. We consider the flow across a series of NACA0012 airfoils \cite{Nigel} with chord length $b=1$, interval $d=1$, and stagger angle $\beta=20^{\circ}$, as shown in Fig. \ref{fig:airfoil}a.
The uniform flow comes from the left in velocity magnitude $1$ with angle of attack $\alpha=10^{\circ}$ (the inflow angle $\gamma=\beta-\alpha=10^{\circ}$). 
The computational domain $\Omega=[-7.5,18.5]\times[-0.5,0.5]$ includes one airfoil placed at $(0,0)$.
The periodic boundary is on the upper and lower sides, the far-field condition ($u=\textup{cos}\gamma,v=-\textup{sin}\gamma$) is on the left and right sides, and the no-slip condition ($u=v=0$) is on the solid wall of the airfoil.
The flow is governed by Eq. (\ref{eq:5}) ($\mu=0.01$), and thus can be predicted by the same pre-trained LNO.
For the present problem, the domain $\Omega$ is extended and divided into $\Omega_1$, $\Omega_2$, and $\Omega_3$ for LNO prediction, as shown in Fig. \ref{fig:airfoil}a (right side).
In each step of time marching shown in Fig. \ref{fig:airfoil}b, firstly, $u_t$ on $\Omega_3$ is obtained by padding. 
The padding size equals to the corrosion width $R(n,N,k)$ of LNO (see  Section \ref{sec5} for detail).
We use constant padding for the far-field BCs and circular padding for the periodic BCs. 
Secondly, LNO takes $u_t$ as the input on $\Omega_1\cup\Omega_2\cup\Omega_3$ and $\tilde{u}_{t+\Delta t}$ as the output on $\Omega_1$.
Thirdly, $\tilde{u}_{t+\Delta t}$ on $\Omega_2$ is obtained using immersed boundary method (IBM) (see Appendix \ref{secA2} for detail).
Thus, $\tilde{u}_{t+\Delta t}$ on the complete domain $\Omega$ (i.e., $\Omega_1\cup\Omega_2$) is obtained by combining the two outputs.

The pre-trained LNO takes the IC (a uniform flow the same as the far-field condition) as the first input and predicts $\{\tilde{u}_{k\Delta t}|k\in\mathbb{N}\}$ recurrently.
We exhibit a series of contours of velocity magnitude for both the steady state (Fig. \ref{fig:airfoil}c) and the transient developing history (Fig. \ref{fig:airfoil}d). 
It is seen that the LNO predicts the flowing patterns accurately. 
Firstly, under the influence of the airfoil, the fluid is “squeezed”, resulting in high-speed regions between the airfoils. 
Behind the airfoil, the flow separation generates a low-speed region, which is slightly biased towards the upper side owing to the positive angle of attack. 
Secondly,  interactions between the periodic airfoils are successfully captured as the periodic BCs are properly introduced.
Thirdly, LNO predicts the evolving history well, which shows a growth of the wake region as $t$ increases and finally reaches a steady state.
Compared to the reference FEM solutions, the mean absolute error is 0.1127 at the steady state. 
Taking the maximum velocity of 1.5 as a reference, the error rate is 7.51\%, which is relatively small. 
Notably, the present method achieves really high efficiency. 
The pre-trained LNO spends 5.96 seconds including boundary treatments on one desktop-level NVIDIA Geforce RTX 2080ti GPU to predict the flow from 0 to 20 seconds, while FEM needs 6124 seconds to do the same on one desktop-level Intel Core i7-7700K CPU. 
It means LNO achieves a speedup of 1027 times for solving the problem. 
Moreover, we also present the results of predicting flows crossing tandem cascades in Fig. \ref{fig:airfoil}e. 
It shows that, despite different boundaries (the airfoils) leading flows to distinct directions, the pre-trained LNO correctly predicts the steady-state streamlines since the governing equations are unchanged.

\section{Discussion}\label{sec4}

This work proposes the local neural operator (LNO) concept to approximate the time-marching operator for transient partial differential equations (PDEs). 
The concept equips LNO with variable computational domains derived from three basic elements, (i) a unit domain of the output functions, (ii) the local-related condition, and (iii) the shift-invariant condition. 
We trained the LNO with randomly generated samples on boundaryless domains. 
Then, the pre-trained LNO collaborates with case-specific boundary treatments to solve the problems governed by the same PDEs on unseen domains. 
As examples, we train the LNO only once to solve different problems, e.g., the free flows on periodic domains, the internal flow in lid-driven cavity, and the external flow across the cascade of airfoils. 
Moreover, the LNO is capable of learning more transient PDEs, e.g., viscous Burgers equation and wave equation, see Appendix \ref{secA3} for detail.

The highlight of the present work is the reusability of the pre-trained LNO on varied computational domains.
The solving process of transient PDEs is decomposed into time marching and boundary imposing. 
With different boundaries imposed independently, one pre-trained LNO is able to solve bunches of physical fields on varied domains, whether it is large, small, regular, or irregular.
Such a broader scope of reuse encourages us to train larger models with more data. 
If many pre-trained models were collected into a library, it would be really convenient for future applications to quickly select proper models and solve various transient PDEs in scientific or engineering scenarios.

The pre-trained LNO realizes great efficiency when being a numerical scheme for solving transient PDEs.
Compared to the conventional implicit schemes, the present LNO models the complex solving procedure as a purely explicit scheme with parallel nature, which brings superior efficiency. 
On the other hand, compared to the conventional explicit schemes, the pre-trained LNO has superior numerical stability to use larger time interval. 
LNO bypasses the limit of numerical stability by learning the time-marching operator of large time interval directly from high-quality samples.
Quantitatively, the Courant-Friedrichs-Lewy (CFL) condition says that an explicit numerical scheme may blow up when its CFL number defined as $\frac{u\Delta t}{\Delta x}$ is greater than the criterion value, which limits the maximum time interval and constrains the lower bound of computational costs.
Usually, the criterion value is lower than 1 for explicit schemes \cite{Tam1993,Cockburn1989}.
The present LNO realized a large CFL number of 3.2 ($\Delta t=0.05,\Delta x=\frac{1}{64},u\approx 1$), allowing a larger time interval to get better efficiency.
To carry forward these advantages of LNO in numerical calculations, the application should be extended to, e.g., the diffusion of mass or heat, the transient deformation of solids, complex dynamics of multi-physics systems, etc.

\section{Methods}\label{sec5}

\begin{figure}[htbp]
    \centering
\includegraphics[width=\textwidth]{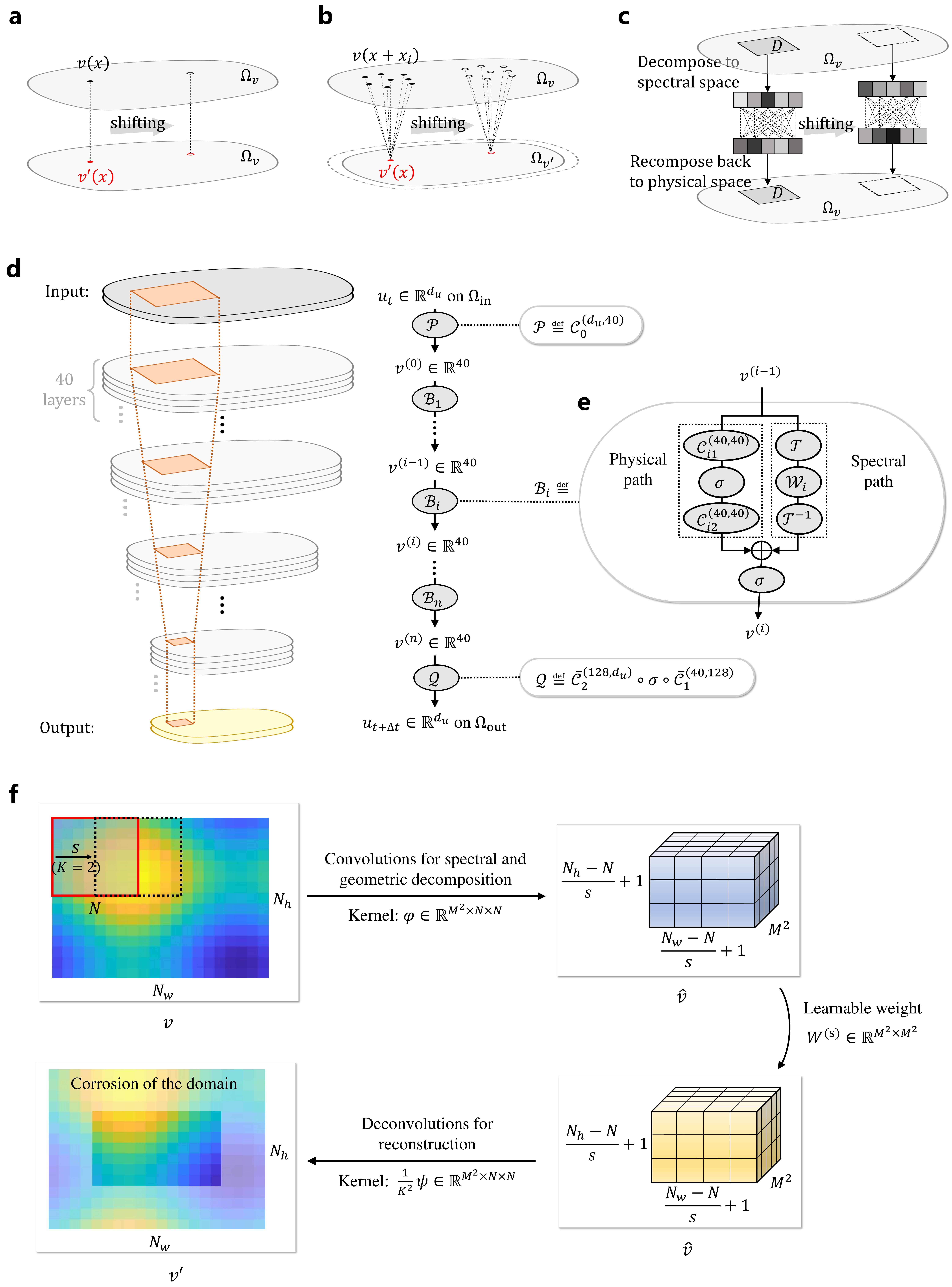}
\caption{\textbf{Technical details in the present LNO architecture.}
\textbf{a}, Pointwise layers.
\textbf{b}, Physical layers.
\textbf{c}, Spectral layers. 
Schematics in a, b and c are basic layers in LNO categorized according to how they link function values at different positions.
\textbf{d}, The present LNO architecture composed using layers in a, b and c.
$\{v^{(i)}\}_{i=0}^n$ are intermediate hidden functions during LNO prediction.
\textbf{e}, The interior blocks. 
\textbf{f}, Implementation of the spectral layers by using discretized convolutions. 
The spectral layer processes a single channel discretized function $v$ on an equidistant grid of size $N_w\times N_h$.
This implementation brings an issue called ‘corrosion of the domain’ caused by insufficient coverage in near-boundary areas.}\label{fig:spectralpath}
\end{figure}

In this section, we introduce technical details of the LNO concept to help better understand and to ensure reliable reproduction of this work.
Primarily, we introduce the specific architecture of LNO, a multi-layer deep neural network comprised of dozens of local-related layers. 
Sequentially, we formulate the local-related layers (Fig. \ref{fig:spectralpath}a, \ref{fig:spectralpath}b, and \ref{fig:spectralpath}c), introduce how they compose the LNO (Fig. \ref{fig:spectralpath}d and \ref{fig:spectralpath}e), and how we conveniently code them using modern open-source deep learning toolkits (Fig. \ref{fig:spectralpath}f). 
After that, the difference between $\Omega_{\textup{in}}$ and $\Omega_{\textup{out}}$ in Eq. (\ref{eq:4}) is discussed based on the LNO architecture implemented.
This “corrosion of the domain” issue is essential in boundary treatment when applying the pre-trained LNOs. 
At last, we provide supplemental details about data generation and LNO training.

\vspace{8mm}

\textbf{Local-related neural operator layers.}
Neural operator layers are basic components of deep neural operators. 
To compose the present LNO, we first introduce three typical neural operator layers and show how they meet the proposed local-related condition in Eq. (\ref{eq:2}).
We categorize the layers according to how they link function values at different positions: (i) no link; (ii) linked directly in physical space; (iii) linked via spectral space.
In following descriptions, the input and the output functions of these layers are scalar functions, and they are respectively denoted as $v(x),x\in\Omega_v$, and $v^{\prime}(x),x\in\Omega_{v^{\prime}}$.

\vspace{8mm}
\emph{I. Pointwise layers.} 
Pointwise layers transform the input function point-by-point independently (Fig. \ref{fig:spectralpath}a). 
For example, the commonly used activation function $\sigma$ in neural networks can be used as a pointwise layer that 
\begin{equation}
\sigma:v(x)\mapsto v^{\prime}(x),\quad \textup{or equivalently,}\quad v^{\prime}(x)=\sigma(v(x)), x\in\Omega_v.
\label{eq:7}
\end{equation}
The formulation of $\sigma$ is optional in practices. We use GELU activation \cite{Hendrycks2016} in this work that
\begin{equation}
\sigma(x)=0.5 x\left(1+\tanh \left[\sqrt{\frac{2}{\pi}}\left(x+0.044715 x^3\right)\right]\right) .
\label{eq:8}
\end{equation}
The domains for $v(x)$ and $v^{\prime}(x)$ are identical for pointwise layers, i.e., $\Omega_{v^{\prime}}=\Omega_v$. 
Since there is no relation between $v(x)$ and $v^{\prime}(x^{\prime})$ while $x\neq x^{\prime}$, the pointwise layers are clearly local.

\vspace{8mm}
\emph{II. Local-related physical layers.} 
The local-related physical layers (denoted as $\mathcal{C}$) approximate a direct relation between function values at different positions in physical space (Fig. \ref{fig:spectralpath}b).
It includes a learnable weight $W^{(\textup{p})}\in\mathbb{R}^I$ with components $W^{(\textup{p})}_i,i=1,2,...,I$. 
The local-related physical layers transform functions,
\begin{gather}
        \mathcal{C}:v(x)\mapsto v^{\prime}(x^{\prime}),\quad x\in\Omega_v,x^{\prime}\in\Omega_{v^{\prime}},\\
    \textup{i.e.,}\quad
    v^{\prime}(x^{\prime})=\mathcal{C}(v(x);W^{(\textup{p})})
    =\sum_{i=1}^IW_i^{(\textup{p})}v(x^{\prime}+x_i),.\nonumber
    \label{eq:9}
\end{gather}
where $x_1,x_2,...,x_I$ are the relative positions of scattered sensors, similar to the concept in DeepONet \cite{LuLu2021}.

It is seen that the maximum related distance between function values is $\max(\left \|x_1 \right \| ,\left \|x_2 \right \|,...,\left \|x_I \right \|)$ which is adjustable.
We design it as limited to ensure the physical layer is local-related. 
Specially, if we let $I=1$ and $x_1=0$ in Eq. (\ref{eq:9}), the layer $\mathcal{C}$ turns into a pointwise layer (denoted as $\bar{\mathcal{C}}$).

\vspace{8mm}
\emph{III. Localized spectral layers.}
Function values at different positions can also be related via spectral space, and meanwhile, the technique of spectral transform can benefit the approximation ability of neural operators \cite{LiZongyi2020}.
The localized spectral layers are shown in Fig. \ref{fig:spectralpath}c. 
We localize the spectral layers by conducting the spectral transform $\mathcal{T}$ on a local subdomain $D\subset\Omega_v$,
\begin{gather}
        \mathcal{T}:v(x)\mapsto \{\hat{v}_m\}_{m=0}^{M^d-1}, \quad x\in D,\\
    \textup{i.e.,}\quad \hat{v}_m=\mathcal{T}(v(x))
    \int_D v(x)\varphi_m(x)dx,\quad m=0,1,...,M^d-1,\nonumber
    \label{eq:10}
\end{gather}
where $\hat{v}_m$ is the $m^{\textup{th}}$ spectral component of $v(x)$, and $\varphi_m(x)$ is the $m^{\textup{th}}$ forward basis function for spectral transform.
After that, we apply a fully connected neural layer on the spectral components. 
Specifically, it is a linear operation $\mathcal{W}$ with a learnable weight matrix $W^{(\textup{s})}\in\mathbb{R}^{M^d\times M^d}$,
\begin{gather}
        \mathcal{W}:\{\hat{v}_m\}_{m=0}^{M^d-1}\mapsto \{\hat{v}^{\prime}_{m^{\prime}}\}_{m^{\prime}=0}^{M^d-1},\\
    \textup{i.e.,}\quad
    \hat{v}^{\prime}_{m^{\prime}}=\mathcal{W}\left(\{\hat{v}_m\}_{m=0}^{M^d-1};W^{(\textup{s})}\right)
    =\sum_{m=0}^{M^d-1}W^{(\textup{s})}_{mm^{\prime}}\hat{v}_m, \quad m^{\prime}=0,1,...,M^d-1.\nonumber
    \label{eq:11}
\end{gather}
where $W^{(\textup{s})}_{mm^{\prime}},m=0,1,...,M^d-1,m^{\prime}=0,1,...,M^d-1$, are components of $W^{(\textup{s})}$).
Finally, we recompose the output components back to the original space by $\mathcal{T}^{-1}$ that
\begin{gather}
        \mathcal{T}^{-1}:\{\hat{v}^{\prime}_{m^{\prime}}\}_{m^{\prime}=0}^{M^d-1}\mapsto v^{\prime}(x),\quad x\in D,\\
    \textup{i.e.,}\quad
    v^{\prime}(x)=\mathcal{T}^{-1}\left(\{\hat{v}^{\prime}_{m^{\prime}}\}_{m^{\prime}=0}^{M^d-1}\right)
    =\sum_{m^{\prime}=0}^{M^d-1}\hat{v}_{m^{\prime}}^{\prime}\psi_{m^{\prime}}(x),\nonumber
    \label{eq:12}
\end{gather}
where $\psi_{m^{\prime}}(x)$ is the $m^{\prime}$-th forward basis function for spectral transform, and $\hat{v}_{m^{\prime}}^{\prime}$ is the $m^{\prime}$-th spectral component of $v^{\prime}(x)$. 
In this work, we derive $\varphi_m(x)$ in Eq. (\ref{eq:10}) and $\psi_{m^{\prime}}(x)$ in Eq. (\ref{eq:12}) by using Legendre polynomials as the spectral basis, which better suit the nonperiodic nature of functions on varied local subdomains than the Fourier polynomials \cite{LiZongyi2020}.
See Appendix \ref{secA4} for more details.
Moreover, we adopt the first $M$ modes at lower frequencies for low-pass filtering.
So far, we obtained the complete spectral layer which is a composition of $\mathcal{T}$ in Eq. (\ref{eq:10}), $\mathcal{W}$ in Eq. (\ref{eq:11}), $\mathcal{T}^{-1}$ in Eq.(\ref{eq:12}), and it transforms $v(x)$ to $v^{\prime}(x)$ on $D$.
By shifting the local subdomain $D$, the spectral layer realizes a mapping on $\Omega_v$.
During the mappings, each two of function values $v(x),x\in D$ and $v^{\prime}(x^{\prime}),x^{\prime}\in D$, are linked by learnable weight. 
It means the maximum related range of this layer depends on $D$ which is adjustable.
We design it as bounded and relatively small to ensure the spectral layers are local-related.

\vspace{8mm}
\emph{Layers with multiple channels.}
The above-introduced three kinds of layers use scalar functions as input and output, yet the physical problems to be solved (especially the multi-physics ones) usually concern vector or/and tensor fields such as the velocity fields in fluids.
We arrange the vector/tensor components into a new dimension, i.e., the channel of the function.
The symbols $v(x)\in\mathbb{R}^{d_v}$ and $v^{\prime}(x)\in\mathbb{R}^{d_{v^{\prime}}}$ means the function is with $d_v$ and $d_{v^{\prime}}$ channels, respectively.
In the form of components, the input and output functions are $\{v_j(x)\}_{j=1}^{d_v}$ and $\{v^{\prime}_{j^{\prime}}(x^{\prime})\}_{j^{\prime}=1}^{d_{v^{\prime}}}$, respectively.
The neural operator layers can transform the multi-channel functions in two ways, i.e., interchange the channels or transform the functions independently in channels.
We interchange the channels in physical layers (including the pointwise physical layers), then Eq. (\ref{eq:9}) changes to 
\begin{equation}
\begin{aligned}
    \mathcal{C}^{\left(d_v, d_{v^{\prime}}\right)}:&\left\{v_j(x)\right\}_{j=1}^{d_v} \mapsto\left\{v_{j^{\prime}}^{\prime}\left(x^{\prime}\right)\right\}_{j^{\prime}=1}^{d_{v^{\prime}}}, \quad x \in \Omega_v, x^{\prime} \in \Omega_{v^{\prime}},\\
    \textup{i.e.,}\quad
    v_{j^{\prime}}^{\prime}\left(x^{\prime}\right)&=\mathcal{C}^{(d_v,d_{v^{\prime}})}\left(\{v_j(x)\}_{j=1}^{d_v};W^{(\textup{p})}\right)\\
    &=\sum_{j=1}^{d_v} \sum_{i=1}^I W_{i j j^{\prime}}^{(\mathrm{p})} v_j\left(x^{\prime}+x_i\right), \quad j^{\prime}=1,2, \ldots, d_{v^{\prime}}.
\end{aligned}
\label{eq:13}
\end{equation}
There appear $d_v\times d_{v^{\prime}}$ links between the input and output functions.
Each link has an independent learnable weight, i.e., compared to the single channel physical layer in Eq. (\ref{eq:9}), the learnable weight is expanded by $d_v\times d_{v^{\prime}}$ times that $W^{(\textup{p})}\in \mathbb{R}^{I\times d_v\times d_{v^{\prime}}}$ and $W_{i j j^{\prime}}^{(\mathrm{p})}$ is the component.
On the contrary, we let pointwise activations and the spectral layers transform the multi-channel functions channel-by-channel independently.
Thus, the number of channels of the input and output functions should be the same ($d_v=d_{v^{\prime}}$). 
In spectral layers, each channel has an independent learnable weight, i.e., compared to the single-channel spectral layer in Eq. (\ref{eq:11}), the learnable weight is expanded by $d_v$ times that $W^{(\textup{s})}\in \mathbb{R}^{M^d\times M^d\times d_v}$.

\vspace{8mm}

\textbf{Local-related neural operator layers compose the architecture of LNO. }
Architecting deep neural networks is a work with huge space for imagination. 
The specific LNO architecture composed in this work shows an example. 

To approximate the operator $\mathcal{G}$ in Eq. (\ref{eq:1}), a lifting-projection structure of neural operators is formulated as \cite{Kovachki2021,LiZongyi2020}
\begin{equation}
\mathcal{G}_\theta \stackrel{\text { def }}{=} \mathcal{Q} \circ \mathcal{B}_n \circ \ldots \circ \mathcal{B}_1 \circ \mathcal{P},
\label{eq:14}
\end{equation}
where $\theta\in\mathbb{R}^{N_{\theta}}$ is the set of all learnable weights in $\mathcal{G}_\theta$, and $N_{\theta}$ is the amount of real-number components in $\theta$.
$\mathcal{P}$ and $\mathcal{Q}$ are the lifting and projection mapping, respectively. 
`$\circ$' is the symbol for composite mappings.
$\{\mathcal{B}_i\}_{i=1}^n$ are the interior mapping blocks with $n$ being the number of blocks.

Herein, we compose LNO on the structure Eq. (\ref{eq:14}) to approximate the local-related operator $\mathcal{G}_L$ in Eq. (\ref{eq:4}), as shown in Fig. \ref{fig:spectralpath}d.
The used components include GELU activation layer $\sigma$ described in Eqs. (\ref{eq:7}-\ref{eq:8}), physical layers $\mathcal{C}$ (includes the pointwise ones $\bar{\mathcal{C}}$) in Eq. (\ref{eq:9}), spectral layers $\mathcal{T}^{-1}\circ\mathcal{W}\circ\mathcal{T}$ in Eqs. (\ref{eq:10}-\ref{eq:12}).
In what follows, we use subscripts to identify layers/operations with independent learnable weights.
Specifically, let $\mathcal{P}\stackrel{\text { def }}{=}\mathcal{C}_0^{(d_u,40)}$, i.e., a physical layer $\mathcal{C}_0^{(d_u,40)}$ lifts the input functions from $d_u$ channels to 40 channels for enriching the capability of representation of LNO. 
At the end of the network, we let $\mathcal{Q}\stackrel{\text { def }}{=}\bar{\mathcal{C}}_2^{(128,d_u)}\circ\sigma\circ\bar{\mathcal{C}}_1^{(40,128)}$, i.e., the two pointwise physical layers project the interior function of 40 channels back to $d_u$ channels to match the output physical fields $u_{t+\Delta t}(x)$.
The lifting and projection blocks in LNO are similar to that in FNO \cite{LiZongyi2020}.

We design the inner blocks $\{\mathcal{B}_i\}_{i=1}^n$ distinctively (Fig. \ref{fig:spectralpath}e) as 
\begin{equation}
\mathcal{B}_i \stackrel{\text { def }}{=} \sigma \circ\left(\mathcal{C}_{i 2}^{(40,40)} \circ \sigma \circ \mathcal{C}_{i 1}^{(40,40)}+\mathcal{T}^{-1} \circ \mathcal{W}_i \circ \mathcal{T}\right), \quad i=1,2, \ldots, n.
\label{eq:15}
\end{equation}
The two terms $\mathcal{C}_{i 2}^{(40,40)} \circ \sigma \circ \mathcal{C}_{i 1}^{(40,40)}$ and $\mathcal{T}^{-1} \circ \mathcal{W}_i \circ \mathcal{T}$ are parallel paths that process the input functions respectively in physical and spectral space. 
The present LNO includes 4 inner blocks ($n=4$). 
In each inner block, the input function is separately processed by the two paths and then added together.
In the entire architecture of LNO, the two paths branch and merge several times, which provides LNO with a highly complex space to better approximate the desired operator $\mathcal{G}_L$ in Eq. (\ref{eq:4}).

\vspace{8mm}

\textbf{A convenient implementation by using discretized convolutions.}
The functions and layers in LNO should be presented in a discretized form for practice. 
Here, we consider an equidistant grid discretization with size $\Delta x$. 
Thus, the layers in LNO can be realized by discretized convolutions widely used in image processing \cite{Lecun1998}. 
These discretized convolutions are convenient to code with deep learning toolkits like PyTorch \cite{Paszke2019} and achieve great computational efficiency on GPUs.
The following descriptions are for 2-D cases, and it is easy to extend to 1-D or 3-D cases.

The only requirement to apply the pointwise layers is that the input and output functions are discretized identically.
Herein, the used discretization of equidistant grids fulfills this requirement.

The physical layers in LNO are implemented by using discretized convolutional neural layers with learnable kernel weights. 
Based on Eq. (\ref{eq:13}), we design sensors $x_1,x_2,...,x_I$ as $\{x_{i_1i_2}=(i_1\Delta x,i_2\Delta x\}_{i_1,i_2=-H}^H$. 
Then, Eq. (\ref{eq:13}) is transformed to 
\begin{equation}
v_{j^{\prime}}^{\prime}\left(x_{k_1 k_2}^{\prime}\right)=\sum_{j=1}^{d_v} \sum_{i_1, i_2=-H}^H W_{i_1 i_2 j j^{\prime}}^{(\mathrm{p})} v_j\left(x_{k_1 k_2}^{\prime}+x_{i_1 i_2}\right), \quad x_{k_1 k_2}^{\prime} \in \Omega_{v^{\prime}, j^{\prime}}=1,2, \ldots, d_{v^{\prime}};
\label{eq:16}
\end{equation}
where $\{x_{k_1 k_2}^{\prime}=(k_1\Delta x,k_2\Delta x)\}_{k_1,k_2\in\mathbb{Z}}$ is the discretized variable.
Eq. (\ref{eq:16}) equals a discretized convolutional layer from $d_v$ to $d_{v^{\prime}}$ channels with stride 1 and $(2H+1)\times(2H+1)$ kernels, and the kernel weight $W^{(\textup{p})}\in\mathbb{R}^{(2H+1)\times(2H+1)\times d_v\times d_{v^{\prime}}}$ is learnable. 
All physical layers in the present LNO architecture use $3\times 3$ kernels, i.e., $H=1$.

The spectral layers can also be realized by discretized convolutions as shown in Fig. \ref{fig:spectralpath}f. 
Firstly, the integration on $D$ in Eq. (\ref{eq:10}) is obtained via Gaussian quadrature. 
The function values on Gaussian points are interpolated from that on the given equidistant point $\{x_{i_1 i_2}=(i_1\Delta x,i_2\Delta x)\}_{i_1,i_2=0}^{N-1},x_{i_1 i_2}\in D$. 
Then, Eq. (\ref{eq:10}) is transformed to
\begin{equation}
\hat{v}_{m, k_1 k_2}=\sum_{i_1=0}^{N-1} \sum_{i_2=0}^{N-1} v\left(x_{k_1 k_2}+x_{i_1 i_2}\right) \varphi_{m, i_1 i_2}, \quad x_{k_1 k_2}+x_{i_1 i_2} \in \Omega_v, m=0,1, \ldots, M^2-1.
\label{eq:17}
\end{equation}
where $\{x_{k_1 k_2}=(k_1s\Delta x,k_2s\Delta x)\}_{k_1,k_2\in\mathbb{Z}}$ is the discretized variable, and $s$ is the shifting unit. 
Then, Eq. (\ref{eq:17}) equals a discretized convolutional layer from 1 to $M^2$ channels with an $N\times N$ constant kernel $\varphi\in\mathbb{R}^{M^2\times N\times N}$, where $\varphi_{m,i_1 i_2}$ is the component and the stride is $s$. 
Secondly, the linear operation with learnable weight in Eq. (\ref{eq:11}) is realized by a convolution with kernel size 1 from $M^2$ to $M^2$ channels,
\begin{equation}
{\hat{v}^{\prime}}_{m^{\prime}, k_1 k_2}=\sum_{m=0}^{M^2-1} W_{m m^{\prime}}^{(\mathrm{s})} \hat{v}_{m, k_1 k_2}, \quad k_1, k_2 \in \mathbb{Z}, m^{\prime}=0,1, \ldots, M^2-1.
\label{eq:18}
\end{equation}
Thirdly, the recombination of models in Eq. (\ref{eq:12}) is realized by using discretized deconvolution formula \cite{Zeiler2011} (also called fractionally-strided convolution) from $M^2$ to 1 channel using constant kernel weight $\psi\in\mathbb{R}^{M^2\times N\times N}$ ($\psi_{m^\prime,i_1 i_2}$ denotes the component) that
\begin{equation}
v^{\prime}(x)=\frac{1}{K^2} \sum_{x_{k_1 k_2}+x_{i_1 i_2}=x} \sum_{m^{\prime}=0}^{M^2-1}{\hat{v}^{\prime}}_{m^{\prime}, k_1 k_2} \psi_{m^{\prime}, i_1 i_2}, \quad x \in \Omega_v,
\label{eq:19}
\end{equation}
where $K=\frac{N}{s}$ is the number of repetitions in one dimension.
In 2-D problems, there are totally $K^2$ combinations of $k_1,k_2,i_1,i_2$ to satisfy the condition $x_{k_1 k_2}+x_{i_1 i_2}=x$ for one given $x$, 
then, we use a normalizing factor $\frac{1}{K^2}$ for dealing with the repetition of output functions caused by shifting.
In Fig. \ref{fig:spectralpath}f, we demonstrate the process of using Eqs. (\ref{eq:17}-\ref{eq:19}), i.e., the implementation of the spectral layers using discretized convolutions. 
In our practices, $M$, $N$, and $K$ of spectral layers are specific for networks and we show them together with the results in Table \ref{tab:1}.
Formulas for $\varphi$, $\psi$, and tables for quick reference of $\varphi_{m,i_1 i_2}$ and $\psi_{m^{\prime},i_1 i_2}$ are in Appendix \ref{secA4}.

\vspace{8mm}

\textbf{Corrosion width of the domain.}
According to Eq. (\ref{eq:4}), LNO predicts the solution functions on $\Omega_{\textup{out}}$ with functions on a bigger domain $\Omega_{\textup{in}}$ as input.
It means that the near-boundary domain $\Omega_{\textup{in}}-\Omega_{\textup{out}}$ is ‘corroded’, thus, we term this issue as ‘corrosion of the domain’. 
The occurrence of domain corrosion is not surprising. 
Some required input values are missing to predict functions on the near-boundary areas by using LNO. 
In other words, the solution operator on these areas is closely related to the BC and quite different from that of the away-boundary domains. 
That is why case-specific treatments are required for the near-boundary areas, and before that, it is necessary to figure out how much of the domain is corroded.

The corrosion width is determined by the architecture. 
It is calculated as follows for the present LNO architecture. 
For pointwise layers, they contribute 0 to the corrosion width. 
Contributions of the physical layers depend on the relative positions of scattered sensors, i.e., $x_1,x_2,...,x_I$ in Eq. (\ref{eq:9}).
In the present implementation by discretized convolutions in Eq. (\ref{eq:16}), a physical layer provides $H\Delta x$, $H=1$, to the corrosion width.
For the spectral layers, according to Eqs. (\ref{eq:17}-\ref{eq:19}), the contribution of the spectral layers to the corrosion width is $\frac{K-1}{K}N\Delta x$, which is derived by considering the related range and the number of repetitions $K=\frac{N}{s}$, as shown in Fig. \ref{fig:spectralpath}f. 
Overall, the corrosion width of the complete architecture is the combination of all the components: 
for series-connected subparts, sum their donations together; 
for parallel-connected subparts, the largest one determines the overall corrosion width. 
Thus, the corrosion width of the present architecture is parameterized by $N,K$ and the number of inner blocks $n$ as $R(n,N,K)$,
 \begin{equation}
\begin{aligned}
R(n, N, k)  =&r_1+n r_2+r_3 \\
= & \Delta x+n \cdot \max \left(\frac{k-1}{k} N, 2\right) \Delta x+0 \\
= & {\left[1+n \cdot \max \left(\frac{k-1}{k} N, 2\right)\right] \Delta x }
\end{aligned}
\label{eq:20}
\end{equation}
where $r_1,r_2,r_3$ are the donations of lifting layers, inner blocks, and projection layers, respectively. 
For the present LNO architecture, the unit input domain ($D_2$) and output domain ($D_1$) are square with the size of $\frac{N}{k}\Delta x+2R$ and $\frac{N}{k}\Delta x$, respectively. 
The minimum local-related range in Eq. (\ref{eq:2}) is then easily obtained, i.e., $r_{\textup{min}}=\frac{N}{k}\Delta x+R$.

\vspace{8mm}

\textbf{Data generation and LNO training.}
For learning N-S equations, the data samples for LNO training and validation are generated using  $Q_2-P_1$ FEM  with implicit Euler scheme \cite{Brezzi1991}. 
The problem is defined on a square domain $[-1,1]\times[-1,1]$ with periodic BC.
The fields start from zero and then are driven by a random external force term
\begin{equation}
F(x, y)=[\sin \pi x \sin 2 \pi x \cos \pi x \cos 2 \pi x] \Lambda[\sin \pi y \sin 2 \pi y \cos \pi y \cos 2 \pi y]^T,
\label{eq:21}
\end{equation}
where $\Lambda =\{\lambda_{ij}\}~(i,j=1\sim 4)$ is a random matrix with $\lambda_{ij}\sim N(0,1)$. 
The external force $F(x,y)$ of $d_u$ channels acts for 0.05 seconds to generate a random velocity field as IC.
The velocity fields are recorded as data samples after removing the external force. 
The training dataset totally contains 1000s of the random flows.
The time interval is $\Delta t=0.05$, i.e., velocity fields at 20000 time levels are recorded.
Moreover, the samples are augmented by coordinate transformations, including rotation and flip.
We adopt 7 different transformations for the 2-D problems: rotate by $90^{\circ}/180^{\circ}/270^{\circ}$, and flip along lines of $x=0$, $y=0$, $y=x$, and $y=-x$.

The data samples are organized by bootstrap for training, i.e., pieces of samples $\{u_{t+k\Delta t}\}_{k=1}^{10}$ are extracted according to a random $t$.
In each iteration, with the randomly sampled $u_t$ as the initial input, the network is trained via decreasing the mean $L_2$ loss between the prediction $\{\tilde{u}_{t+k\Delta t}\}_{k=1}^{10}$ and the real solution $\{u_{t+k\Delta t}\}_{k=1}^{10}$ defined as
\begin{equation}
\mathcal{L}=\frac{1}{10} \sum_{k=1}^{10}\left\|u_{t+k \Delta t}-\tilde{u}_{t+k \Delta t}\right\|_2,
\label{eq:22}
\end{equation}
where $u_{t}$ and $\tilde{u}_{t}$ are in a discretized form of being in $\mathbb{R}^{2\times 128^2}$ that an equidistant grid with spacing $\Delta x=1/64$ is adopted, i.e., the total number of nodes is $128^2$.
All the networks in this work are trained following the same schedule of 100k iterations. 
The optimizer used is Adam  \cite{Kingma2015}. The initial learning rate is set as 0.001 and is manually multiplied by 0.7 every 10k iterations.

\backmatter

\section*{Funding}
This work was supported by NSFC (No.52176043).

\section*{Code availability}
The code accompanying this paper is available in GitHub at \href{https://github.com/PPhub-hy/torch-local-neural-operators}{https://github.com/PPhub-hy/torch-local-neural-operators}.

\bibliography{library}

\begin{appendices}
\newpage

\section{Finite element method (FEM) for reference solutions}\label{secA1}
Here introduces the FEM numerical method used in this work to obtain data samples and reference solutions. 
The numerical method is comprised of schemes for spatial discretization and time marching. 
Specifically, the linear finite element method (FEM) is adopted for spatial discretization for the viscous Burgers equation and wave equation, while $Q_2-P_1$ FEM \cite{Brezzi1991} is used for N-S equations. 
About the time marching schemes, we use implicit Euler for N-S equations and viscous Burgers equations, and implicit Newmark scheme \cite{Gravouil2001} for wave equations. 

\vspace{8mm}

\textbf{$Q_2-P_1$ scheme for N-S equations.}
The main difficulty in the numerical discretization of incompressible N-S equations is to handle the coupling of velocity and pressure. 
When the velocity $u$ and pressure $p$ are discretized using the same polynomial order in FEM, spurious pressure modes occur because the compatibility condition of velocity and pressure (also called inf-sup condition or Ladyzhenškaya-Babuška-Brezzi condition) \cite{Sani1981} is not satisfied. 
Herein, this problem is solved by the staggered grid, such as the $Q_2-P_1$ scheme of FEM \cite{Deville2004}.
Concretely, $u$ is approximated by quadratic polynomials, and $p$ is approximated by linear polynomials.
It consequently requires more nodes for velocities resulting in a node number of approximately $2^d$ times that of pressure, where $d$ is the number of dimensions.
Compared to those schemes which approximate $u$ and $p$ using the same order, for example, the projection method \cite{Guermond2006} or the least-square method \cite{Pontaza2003}, $Q_2-P_1$ scheme requires no auxiliary variables or additional hypotheses, hence can provide us reliable reference solutions.

$Q_2-P_1$ FEM is used to discretize the incompressible N-S equations together with implicit Euler temporal scheme.
The time step $\Delta t$ is 0.05. 
FThe discretized N-S equations are solved by Newton-Raphson iteration, which stops until the maximum increment of velocities is smaller than $10^{-7}$.

 \vspace{8mm}

\textbf{Detailed information for solving flow in the lid-driven cavity.}
The cavity of $[0,3]\times[0,3]$ is segmented into $96^2$ uniform square elements, thus there are $193^2$ nodes for velocities and $97^2$ points for pressure. 
When predicting $u_{t+\Delta t}$ with the present LNO, it takes values on the whole computational domain $\Omega$ as input and predicts the solution function on the away-boundary area $\Omega_1$.
$\Omega_1$ is with $71^2$ square elements and $143^2$ nodes which are determined by the width of corrosion that $R(n,N,k)=25$, and $143=193-25\times2$.
Accordingly, the near-boundary area $\Omega_2$ is with nodes $193^2-143^2=16800$. 
$u_{t+\Delta t}$ in $\Omega_2$ is calculated by $Q_2-P_1$ FEM with the boundary conditions on the interface obtained from LNO and the solid wall boundary conditions outside.

\vspace{8mm}

\begin{figure}[htbp]%
\centering
\includegraphics[width=0.9\textwidth]{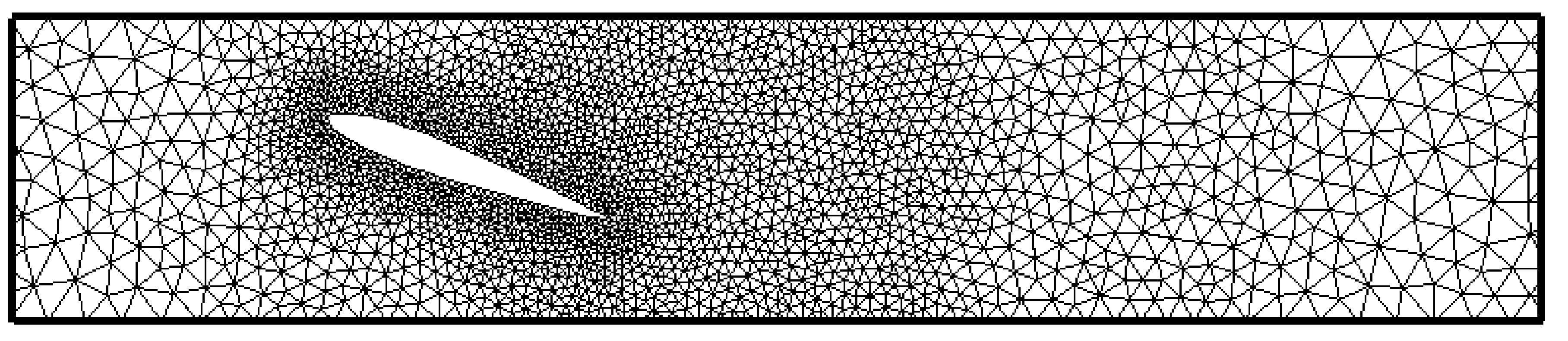}
\caption{The element distribution for reference solution by FEM in solving flow across the cascade of airfoils.}\label{fig:airfoilmesh}
\end{figure}

\textbf{Detailed information for solving flow across the cascade of airfoils.}
The computational domain $[-7.5,18.5]\times[-0.5,0.5]$ is discretized with triangular elements to get the reference solution of this problem by FEM.
We use the locally refined strategy for spatial discretization which is an inborn advantage of FEM with unstructured discretization and is commonly used in practices and makes the calculation efficient. 
Fig. \ref{fig:airfoilmesh} presents the concrete mesh of this problem, it is seen that elements are smaller near the airfoil and in the wake region to  describe the flow separation better, while the mesh is coarse for the outer parts. 
As the prediction of LNO is based on a structured mesh with size $\Delta x=1⁄64$, the mesh for FEM in Fig. \ref{fig:airfoilmesh} is generated by setting the minimum mesh size around $1⁄32$ (therefore, the intervals between nodes of velocities are $1⁄64$) to make the comparison of time consumption reasonable. 
There are totally 7944 elements, 15987 nodes for velocities, and 4021 nodes for pressure.

\newpage
\section{Immersed boundary method (IBM) for solid wall conditions on airfoils}\label{secA2}

\begin{figure}[htbp]%
\centering
\includegraphics[width=0.5\textwidth]{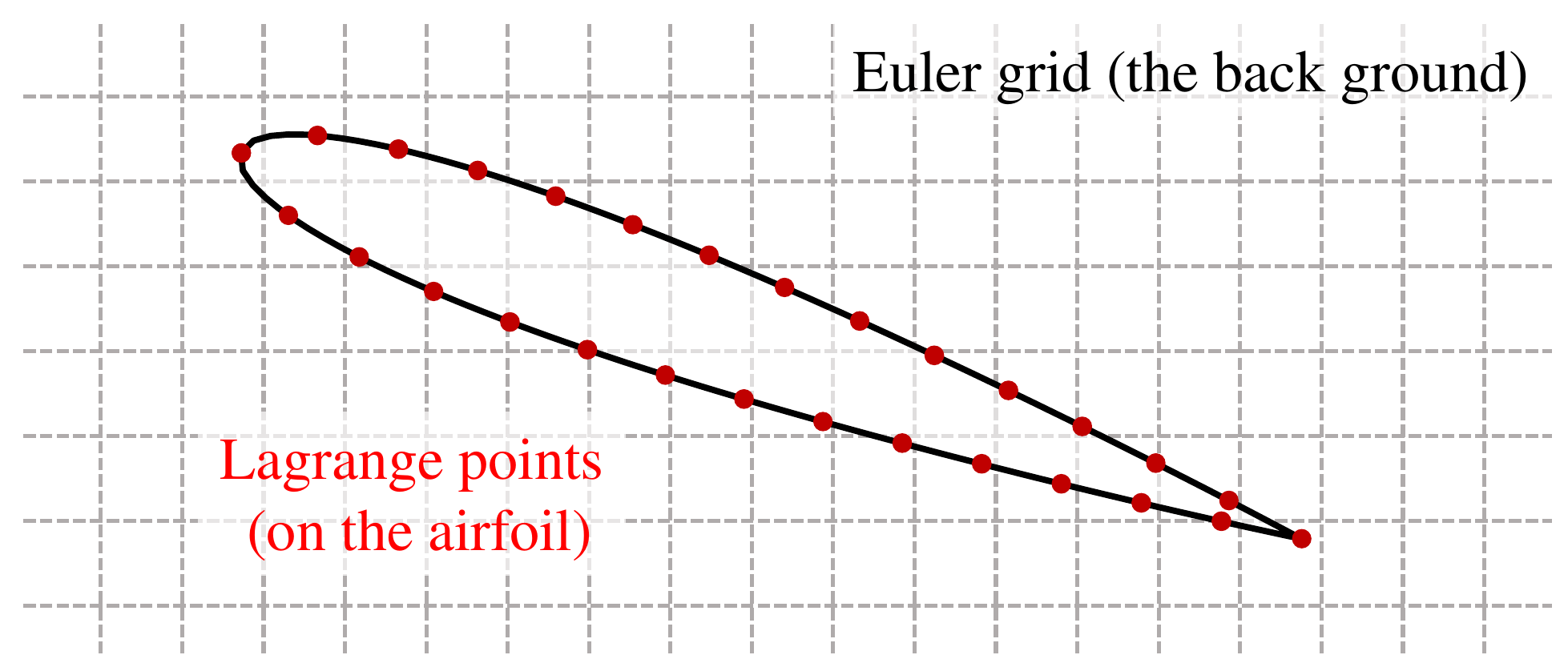}
\caption{Euler points and Lagrange points of the immersed boundary method.}\label{fig:IBM}
\end{figure}

In this work, LNO predicts the velocity fields on equidistant Cartesian grids that do not directly enable treatment on curve boundaries such as the solid wall conditions on airfoils here. 
Fig. \ref{fig:IBM} shows a sketch map for the mesh of the present case, in which the grid points around the obstacle do not fall on the solid wall boundary. 
We address this issue using the immersed boundary method (IBM) \cite{Peskin2002,Uhlmann2005}, by which the effect of boundaries on fluids is converted to treatments on the Cartesian grid point near boundaries.
Thus, the effect of complex curved BC is equivalently imposed.

Concretely, a classic IBM in a direct forcing form is used.
The effect of the solid wall is converted into an external body force, which is imposed via a velocity correction $\Delta u$ on the adjacent grid points. 
The specific values of $\Delta u$ are derived by satisfying the no-slip condition. 
The concrete steps for implementationin a 2-D problem are as follows.
First, predict an intermediate velocity $u^*$ on the Cartesian grid points $x_i$ (called Euler points) by LNO without the solid wall boundary.
Then, interpolate $u^*$ to the points on the airfoil curve $X_j$ (called Lagrange points): 
\begin{equation}
U^*(X_j)=\sum_{i=1}^{N_G} u^*\left(x_i\right) \delta_h\left(x_i-X_j\right)\Delta x^2,\qquad j=1,2,...,N_G^{\text{Lagrange}}
\label{eq:s1}
\end{equation}
where $U^*$ denotes the intermediate velocity on the Lagrange points; 
$N_G$ is the total number of Euler points; 
$N_G^{\text{Lagrange}}$ is the total number of the Lagrange points;
$\Delta x$ is the size of the Cartesian grid;
$\delta_h$ is an approximated delta function, in this paper the 4-point piecewise function \cite{Peskin2002} is applied:
\begin{equation}
\delta_h(x-X)=\frac{1}{\Delta x^2} d\left(\frac{x-X}{\Delta x}\right)
\label{eq:s2}
\end{equation}
with
\begin{equation}
d(r)=w(r_1)w(r_2)
\label{eq:s3new}
\end{equation}
\begin{equation}
w(r_i)=\left\{\begin{array}{c}
\frac{1}{8}\left(3-2|r_i|+\sqrt{1+4|r_i|-4 r_i^2}\right),|r_i|<1, \\
\frac{1}{8}\left(5-2|r_i|-\sqrt{-7+12|r_i|-4 r_i^2}\right), 1 \leq|r_i|<2, \\
0,|r_i| \geq 2 ,
\end{array}\right.
\quad i=1,2,
\label{eq:s3}
\end{equation}
where $r\in \mathbb{R}^2$ and $r_1, r_2$ are components of $r$. 

Next, calculate the volume force $F$ using the boundary condition and interpolate back to the Euler points:
\begin{equation}
F(X)=\frac{U_{BC}(X)-U^*(X)}{\Delta t},
\label{eq:s4}
\end{equation}

\begin{equation}
f(x_i)=\sum_{j=1}^{N_G^{\text{Lagrange}}}F(X_j)\delta_h(x_i-X_j)\Delta x\Delta s,
\label{eq:s5}
\end{equation}
where $\Delta s$ is the interval between Lagrange points.
Finally, we obtain the modified velocity considering the boundary as
\begin{equation}
\Delta u(x_i)=f(x_i)\Delta t,
\label{eq:s6}
\end{equation}

\begin{equation}
u_{t+\Delta t}(x_i)=u^*(x_i)+f(x_i)\Delta t.
\label{eq:s7}
\end{equation}

Briefly speaking, when we use LNO to predict the velocity field $\tilde{u}_{t+\Delta t}$ around irregular objects, first predict the intermediate velocity $u^*$ by LNO with $u_t$ as input, then calculate the velocity correction $\Delta u$ following Eqs. (\ref{eq:s1}-\ref{eq:s6}), finally impose the correction to obtain $u_{t+\Delta t}$ by Eq. (\ref{eq:s7}).

\newpage

\section{Learn different equations with LNO}\label{secA3}

This section briefly demonstrates the universality of LNO to learn various transient PDEs. 
Two fundamental equations derived from physics are considered. 
One is the viscous Burgers equation, which describes the convection and diffusion of physical fields. 
This equation usually generates shocks with sharp gradients in the fields, posing a challenge to numerical solvers including the LNO. 
The other is the wave equation, which appears in acoustics and electromagnetics. 
It is used as a representative problem here for LNO to show the capability of solving second-order transient systems. 
In what follows, we introduce the problem settings and data generation for the two equations. 
Training parameters for these two equations are identical to learning N-S equations in  Section \ref{sec5} of the main text.

\vspace{8mm}

\textbf{Learn viscous Burgers equation. }
The viscous Burgers equation is
\begin{equation}
\frac{\partial u}{\partial t}+u\cdot \nabla  u=\mu\Delta u,
\label{eq:s8}
\end{equation}
where $u$ is the velocity to be solved, $\mu$ is the viscosity. 
LNOs are separately built and trained to learn Eq. (\ref{eq:s8}) with $\mu=0.01$ for 1-D and 2-D cases. 
The LNO takes the velocity function $u_t$ as input and then output $u_{t+\Delta t}$ ($u$ has 1 channel for 1-D case and 2 channels for 2-D case). 

For 1-D case the model problem for training is defined on $[-1,1]$ with periodic boundary condition. 
The initial condition is randomized by
\begin{equation}
u_0(x)=\lambda_1 \sin \pi x+\lambda_2\sin 2\pi x+\lambda_3 \cos \pi x+\lambda_4 \cos 2\pi x,
\label{eq:s9}
\end{equation}
where $\lambda_i\sim N\left(0,1\right)$, $i=1,2,3,4$. 
For 2-D case the problem is defined on a square domain $\left[-1,1\right]\times[-1,1]$ with periodic boundary condition. 
The initial condition $u_0$ is set as $F$ randomized by Eq. (\ref{eq:21}) in the main text. 
We adopt the linear FEM with implicit Euler scheme to generate data samples. 
Equidistant mesh with spacing $\Delta x=1/64$ is used.
The time step is $\Delta t=0.05$.

\vspace{8mm}

\textbf{Learn wave equation.}
The wave equation is 
\begin{equation}
\frac{\partial^2 p}{\partial t^2}-a_0^2 \Delta p=0,
\label{eq:s10}
\end{equation}
where $a_0$ is the velocity of wave propagation, $p$ is the field to be solved. 
LNO is trained to learn Eq. (\ref{eq:s10}) with $a_0=1$.
For this second-order time system, i.e., the highest order of partial derivative with respect to time $t$ is second order, LNO takes $\{p,\frac{\partial p}{\partial t}\}_t$ as the input,  where $p$ and $\frac{\partial p}{\partial t}$ are concatenated in channels. 
The output is $\{p,\frac{\partial p}{\partial t}\}_{t+\Delta t}$ and it is recurrently served as the input for the next step of time marching.

The model problem for training is defined in a square domain $\left[-1,1\right]\times[-1,1]$ with periodic boundary conditions. 
The initial condition $p_0$ is set as $F$ randomized by Eq. (\ref{eq:21}) in the main text, and the initial condition for $\frac{\partial p}{\partial t}$ is set as zero.
We use the implicit Newmark scheme for time marching, which differs from that for N-S equations and Burgers equation as the wave equation is a second-order transient system.
 The implicit Newmark scheme is unconditionally stable \cite{Gladwell1980} and finely suits our data generation. 
Readers please refer to the literature \cite{Bely,Ye2020} for details about the implementation and the principle of parameter selection of implicit Newmark scheme. 
We use an equidistant mesh with spacing $\Delta x=1/64$ and the time step is $\Delta t=0.1$.

\vspace{8mm}

\begin{figure}[htbp]%
\centering
\includegraphics[width=0.9\textwidth]{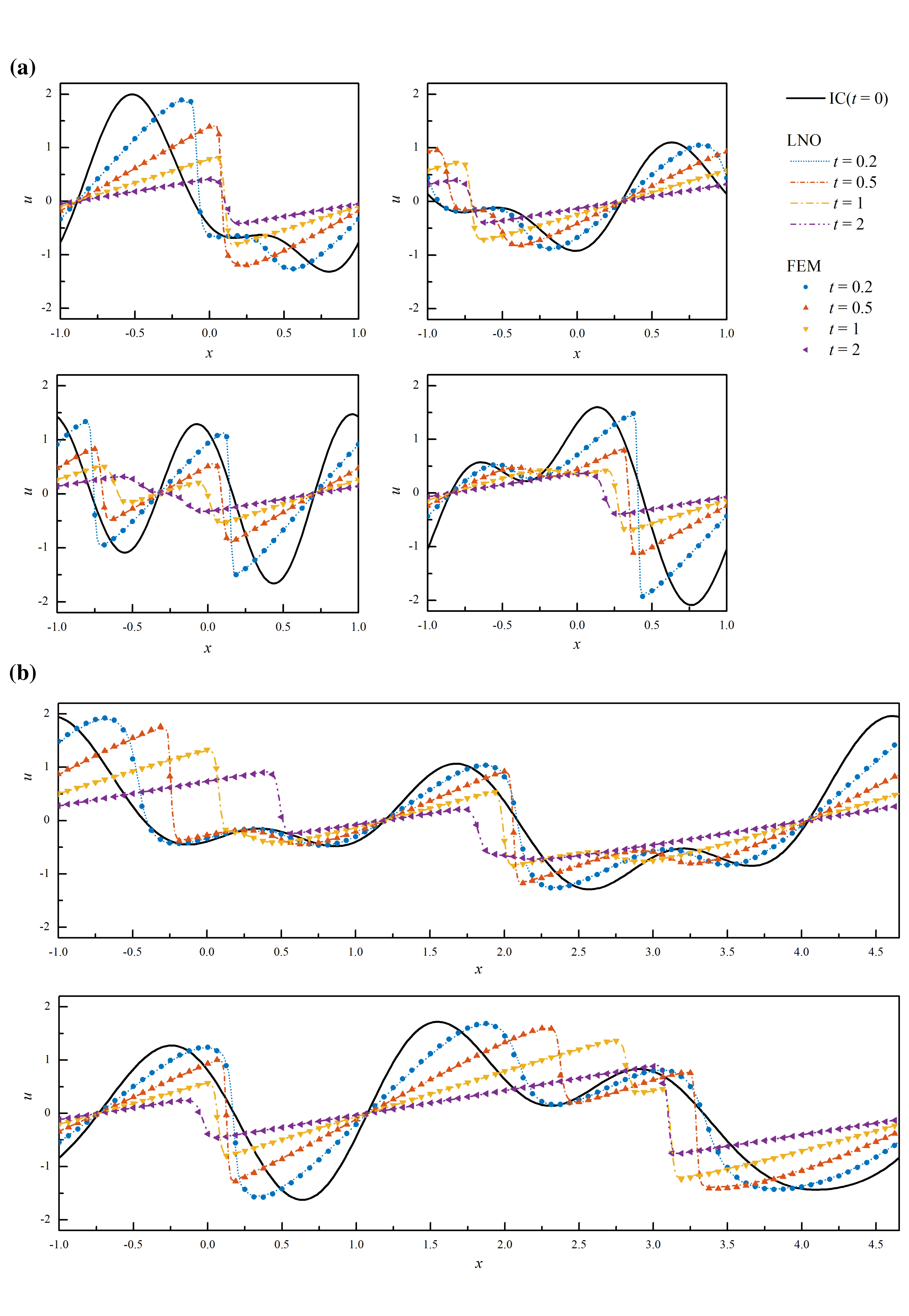}
\caption{\textbf{Results of LNO in solving 1-D viscous Burgers equation (the viscosity $\mu =0.01$).}
Predictions by LNO of randomly generated initial conditions (the solid black lines) are shown in the six figures respectively. 
\textbf{a}, Cases defined in $[-1,1]$. 
\textbf{b}, Cases defined in $[-1,4.656]$. 
All the six results are from one trained LNO. 
Results from FEM are also presented for reference. }
\label{fig:burgers1d}
\end{figure}

\begin{landscape}
\begin{figure}[htbp]%
\centering
\includegraphics[width=1.2\textheight]{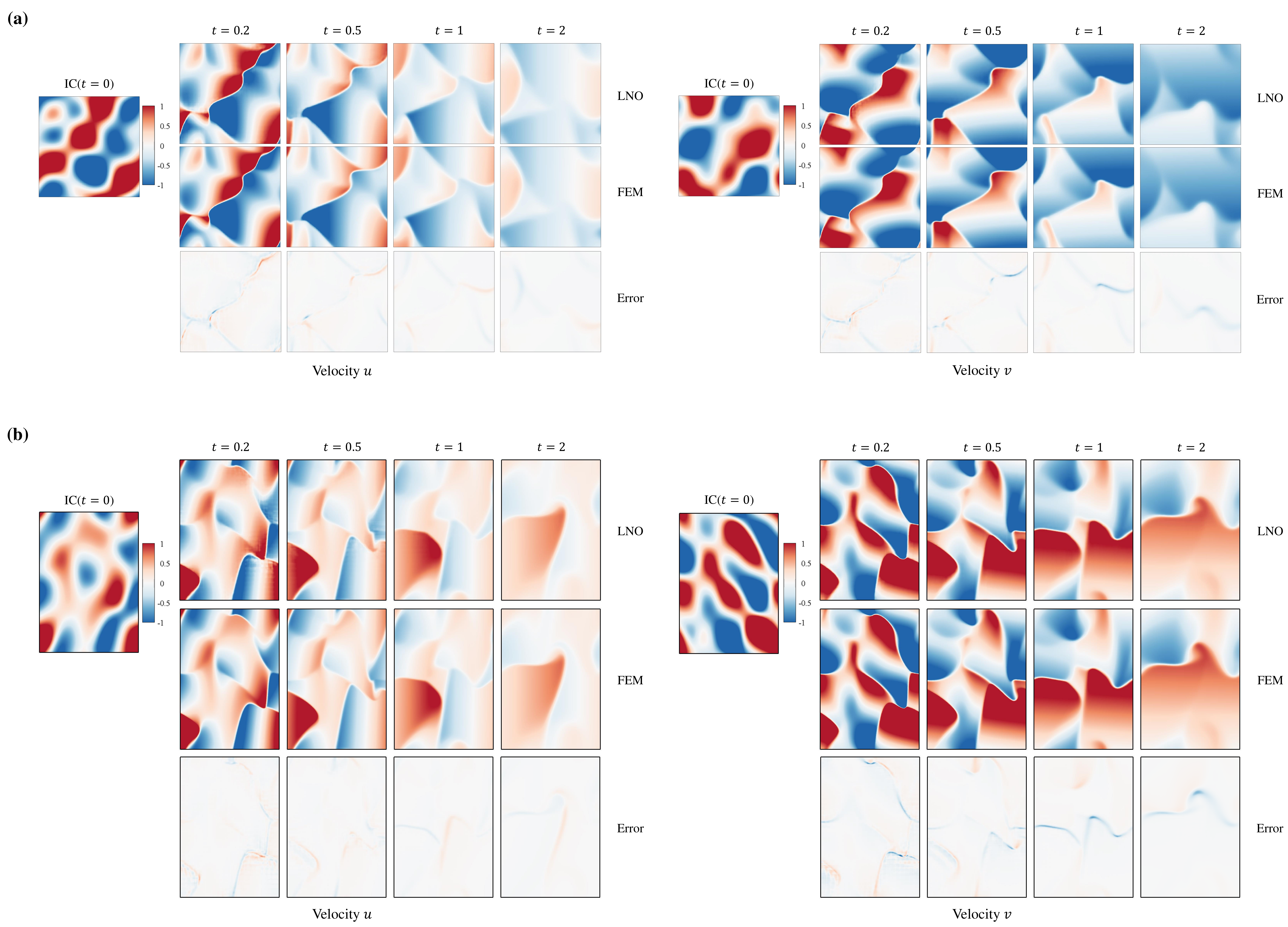}
\caption{\textbf{ Results of LNO in solving 2-D viscous Burgers equations (the viscosity $\mu=0.01$).}
These fields of velocities are predicted by LNO with randomly generated initial conditions as the first input.
\textbf{a}, The case defined in $[-1,1]\times [-1,1]$. 
\textbf{b}, The case defined in $[-1,1]\times [-1,1.828]$.
Results of the two cases are from one trained LNO. 
Results from FEM are also presented for reference.}
\label{fig:burgers2d}
\end{figure}
\end{landscape}

\begin{figure}[htbp]%
\centering
\includegraphics[width=0.9\textwidth]{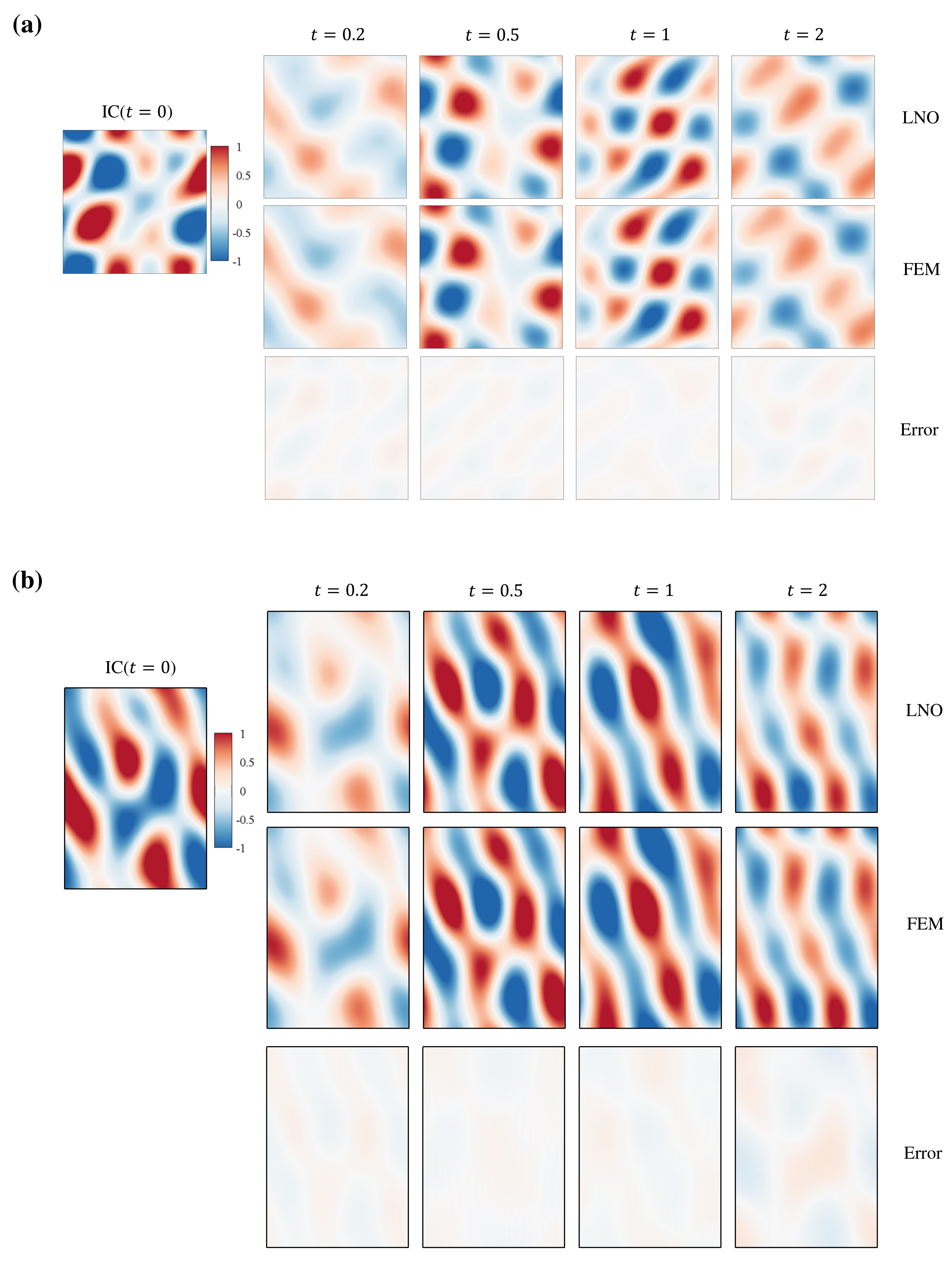}
\caption{\textbf{Results of LNO in solving 2-D wave equation (the wave velocity $a_0=1$).}
These fields of pressure $p$ are predicted by LNO with randomly generated initial conditions as the first input. 
\textbf{a}, The case defined in $[-1,1]\times [-1,1]$. 
\textbf{b}, The case defined in $[-1,1]\times [-1,1.828]$. 
Results of the two cases are from one trained LNO. Results from FEM are also presented for reference.}
\label{fig:wave}
\end{figure}

\textbf{Results.}
The performance of LNO in learning the two equations are measured by the mean $L_2$ error defined in Eq. (\ref{eq:6}) in the main text.
The trained LNOs are examined in predicting $\tilde{u}_{t+\Delta t}$ (or $\tilde{p}_{t+\Delta t}$ for the wave equation) recurrently until $t=2\text{s}$ according to 10 random initial conditions that differ from any training sample.
We compare the results to FNO which is trained and validated following the identical schedule with LNO. 
The key parameters, the number of learnable weights, and the error are listed in Table \ref{tab:s1}. 
These results show that LNO predicts the velocity functions accurately, the error is relatively small and comparable with FNO.
For presenting the results intuitively, we depict the contours of solutions for the three problems respectively in Fig. \ref{fig:burgers1d}, \ref{fig:burgers2d}, and \ref{fig:wave}. 
Each figure includes two groups of cases defined in different computational domains to show that one trained LNO can solve problems in different domains. 
All the cases are with periodic boundary. 
In Figs. \ref{fig:burgers1d}  and \ref{fig:burgers2d} governed by viscous Burgers equations, there are sharp gradients in the predicted fields, which poses a serious challenge to the stability of LNO in the recurrent prediction process. 
In Fig. \ref{fig:wave} governed by the wave equation, the fields are not evolving to a uniform state because of the non-diffusion nature of the equation. 
Commendably, the trained LNO succeeds in predicting these physical fields in domains with diverse shapes with high accuracy maintained, referring to the FEM results shown together with the results by LNO.

\begin{sidewaystable}
\renewcommand\arraystretch{2}
\begin{center}
\begin{minipage}{\textheight}
\caption{Comparison of the mean $L_2$ error between LNO and FNO in solving viscous Burgers and wave equations. 
The averaged error is shown together with the standard deviation of 10 runs.\label{tab:s1}}
\begin{tabular*}{\textheight}{@{\extracolsep{\fill}}cccccccc@{\extracolsep{\fill}}}

\toprule
\multirow{2}{*}{PDE}                      & \multirow{2}{*}{Network} & \multirow{2}{*}{\tnote{*}Parameters}                & \multirow{2}{*}{\makecell{\tnote{**}Number of\\trainable\\weights}} & \multicolumn{4}{c}{$E_t$ (mean $L_2$ error at time $t$)}     \\\cmidrule{5-8}
                                          &                        &                                            &                                                & 0.2s   & 0.5s     & 1s     & 2s     \\
\midrule
\multirow{2}{*}{1-D Burgers ($\mu=0.01$)} & FNO \cite{LiZongyi2020}          & $r_{\text{min}}=\infty$                       &  42957   & 
                      0.050$\pm$0.004  & 0.020$\pm$0.002  & 0.011$\pm$0.002  & 0.008$\pm$0.003 \\\cmidrule{2-8}
                      & The present LNO     &  \makecell{$N=12,M=6,$\\$k=2,r_{\text{min}}=31\Delta x$ }       &  15228   & 0.036$\pm$0.003  & 0.015$\pm$0.002  & 0.007$\pm$0.002  & 0.007$\pm$0.002 \\
\midrule
\multirow{2}{*}{2-D Burgers ($\mu=0.01$)}& FNO \cite{LiZongyi2020}           & $r_{\text{min}}=\infty$                       & 926326     & 
                      0.099$\pm$0.003  & 0.080$\pm$0.005  & 0.070$\pm$0.006  & 0.074$\pm$0.010 \\\cmidrule{2-8}
                      & The present LNO    &   \makecell{$N=16,M=8,$\\$k=2,r_{\text{min}}=41\Delta x$}        & 328656     & 0.055$\pm$0.001  & 0.034$\pm$0.002  & 0.028$\pm$0.003  & 0.031$\pm$0.004 \\
\midrule
\multirow{2}{*}{Wave ($a_0=1$)}& FNO \cite{LiZongyi2020}           & $r_{\text{min}}=\infty$                                & 926326     & 
                      0.053$\pm$0.001  & 0.040$\pm$0.001  & 0.037$\pm$0.001  & 0.049$\pm$0.002 \\\cmidrule{2-8}
                      & The present LNO    &  \makecell{$N=24,M=8,$\\$k=2,r_{\text{min}}=61\Delta x$ }        & 162128     & 0.052$\pm$0.001  & 0.039$\pm$0.001  & 0.036$\pm$0.001  & 0.048$\pm$0.003 \\
\bottomrule
\end{tabular*}
\begin{tablenotes}
\item * $N,M,k$ are the window size, the number of adopted modes, and the number of repetitions, respectively.
$r_{\text{min}}=\frac{N}{k}\Delta x+R(n,N,k)$ is the local-related range. $\Delta x=1/64$. See Section \ref{sec5} for $R(n,N,k)$ and more details.
\item ** The complex weights of FNO are counted twice.
\end{tablenotes}
\end{minipage}
\end{center}
\end{sidewaystable}

\newpage
\section{Legendre spectral transforms for the spectral layers}\label{secA4}

In the spectral path of LNO, we use Legendre polynomials as the basis for the spectral transform \cite{Shen2011} to suit the nonperiodic nature of functions on arbitrary local subdomains.
These polynomials are constructed by carrying out Gram-Schmidt orthogonalization on the polynomial basis $\{1,x,x^2,\ldots,x^n,\ldots\}$ that
\begin{equation}
L_m(x)=\sum_{l=0}^{\left[\frac{m}{2}\right]}(-1)^l \frac{(2 m-2 l) !}{2^m l !(m-l) !(m-2 l) !} x^{m-2 l},
\label{eq:s11}
\end{equation}
where $L_m$ denote $m^{\text{th}}$ Legendre polynomial.

With Legendre polynomials equipped, a 1-D continuous function $f(x)$ defined on $\left[-1,1\right]$ can be approximated by:
\begin{equation}
f(x) \approx \sum_{m=0}^{N-1} \hat{f}_m L_m(x),
\label{eq:s12}
\end{equation}
with
\begin{equation}
\hat{f}_m=\frac{\left(f(x), L_m(x)\right)}{\left(L_m(x), L_m(x)\right)}=\int_{-1}^1 f(x) \frac{L_m(x)}{\left(L_m(x), L_m(x)\right)} d x,
\label{eq:s13}
\end{equation}
where ${\hat{f}}_m$ is the component of $m^{\text{th}}$ mode;
$\left(\cdot,\cdot\right)$ denotes the inner product with weight 1 on $\left[-1,1\right]$.

To derive the normalized basis $\varphi$ and $\psi$ used in spectral layers, first the random local domain D in Eqs. (\ref{eq:10}-\ref{eq:12}) is mapped to the reference element (here is $\left[-1,1\right]$). 
Then by comparing Eq. (\ref{eq:10}) and Eq. (\ref{eq:s13}), the normalized basis of the spectral transform is:
\begin{equation}
\varphi_m(x)=\frac{L_m(x)}{\left(L_m(x), L_m(x)\right)} .
\label{eq:s14}
\end{equation}
Here the mapping coefficient from $D$ to the reference element is ignored for convenience as it is constant for domain $D$ of certain size.
Similarly, the normalized basis of the inverse transform in Eq. (\ref{eq:12}) is as follows:
\begin{equation}
\psi_m(x)=L_m(x) .
\label{eq:s15}
\end{equation}

When the 1-D function $f(x)$ is discretely given, a spectral transform in the discrete form is required. 
The inner product $\left(\cdot,\cdot\right)$ in Eq. (\ref{eq:s13}) is replaced by Gauss quadrature as:
\begin{equation}
\left(f(x), L_{m}(x)\right)=\int_{-1}^{1} f(x) L_{m}(x) d x \approx \sum_{k=0}^{N-1} \omega_{k} f\left(x_{k}\right) L_{m}\left(x_{k}\right), 
\label{eq:s16}
\end{equation}

\begin{equation}
\left(L_{m}(x), L_{m}(x)\right)=\int_{-1}^{1}\left[L_{m}(x)\right]^{2} d x \approx \sum_{k=0}^{N-1} \omega_{k}\left[L_{m}\left(x_{k}\right)\right]^{2},
\label{eq:s17}
\end{equation}
where $\left\{x_k,\omega_k\right\}_{k=0}^{N-1}$ denotes ${(N-1)}^{\text{th}}$-order Legendre-Gauss-Lobatto (LGL) quadrature nodes and weights \cite{Shen2011}. 
$\left\{x_k\right\}_{k=0}^{N-1}$ are the zeros of $\left(1-x^2\right){L^\prime}_{N-1}(x)$ with no explicit expressions, which is usually computed by numerical approaches.
Then, the weights $\left\{\omega_k\right\}_{k=0}^{N-1}$ can be calculated explicitly by
\begin{equation}
\omega_{k}=\frac{2}{N(N-1)\left[L_{N-1}\left(x_{k}\right)\right]^{2}}
\label{eq:s18}
\end{equation}

For practice, when a 1-D function $f(x)$ is discretely given at a series of points $\{{\widetilde{x}}_i\}_{i=0}^{\widetilde{N}-1}$, first we should map these points from D to the reference element (for 1-D case $[-1,1]$ and for 2-D case $[-1,1]\times[-1,1]$).
However, similar to Eq. (\ref{eq:s14}), in our practice the mapping coefficient is ignored as it is constant for domain $D$ of certain size.
Then, interpolate $f({\widetilde{x}}_i)$ to LGL points $\{{x_k}\}_{k=0}^{N-1}$:
\begin{equation}
f\left(x_k\right)=\sum_{i=0}^{\tilde{N}-1} a_{k i} f\left(\tilde{x}_i\right)
\label{eq:s19}
\end{equation}
here the coefficients $a_{ki}$ depend on the interpolation order selected. 
Then, replace the inner product $\left(\cdot,\cdot\right)$ in Eq. (\ref{eq:s13}) by the discrete inner product with ${(N-1)}^{\text{th}}$-order LGL quadrature:
\begin{equation}
\hat{f}_m=\frac{\sum_{k=0}^{N-1} \omega_k f\left(x_k\right) L_m\left(x_k\right)}{\sum_{k=0}^{N-1} \omega_k L_m{ }^2\left(x_k\right)}=\frac{\sum_{k=0}^{N-1} \sum_{i=0}^{\tilde{N}-1} \omega_k a_{k i} f\left(\tilde{x}_i\right) L_m\left(x_k\right)}{\sum_{k=0}^{N-1} \omega_k L_m{ }^2\left(x_k\right)},
\label{eq:s20}
\end{equation}
Then the discrete normalized basis is:
\begin{equation}
\varphi_{m,i}=\frac{\sum_{k=0}^{N-1} \omega_k a_{k i} L_m\left(x_k\right)}{\sum_{k=0}^{N-1} \omega_k L_m^2\left(x_k\right)}.
\label{eq:s21}
\end{equation}
The discrete normalized basis of the inverse transform is:
\begin{equation}
\psi_{m, i}=L_m\left(\tilde{x}_i\right).
\label{eq:s22}
\end{equation}

For 2-D problems, the normalized basis can be obtained by the product of 1-D bases $\varphi_{m,i}$ and $\psi_{m,i}$ with respect to $x$ and $y$ axes that
\begin{equation}
\varphi_{m, i_1 i_2}=\varphi_{p, i_1} \varphi_{q, i_2},
\label{eq:s23}
\end{equation}
\begin{equation}
\psi_{m, i_1 i_2}=\psi_{p, i_1} \psi_{q, i_2},
\label{eq:s24}
\end{equation}
where $m=m(p,q)$.

In this paper, we concretely set $\widetilde{N}=N$, ${\widetilde{x}}_i$ be equidistant points, and the interpolation from $f\left({\widetilde{x}}_i\right) $to $f\left(x_k\right)$ in Eq.(\ref{eq:s19}) be linear.
With these settings, values of $\varphi_{m,i}$ and $\psi_{m,i}$ used in this paper are computed by Eqs. (\ref{eq:s21}-\ref{eq:s22}) and listed in Tables \ref{tab:s2}, \ref{tab:s3}, \ref{tab:s4} for $N=12,18,24$, respectively.

\begin{table}[h]

\begin{subtable}[h]{\linewidth}
\centering

\begin{tabular}{ccccccccc}

\toprule
$\varphi_{m,i}$ & $m=1$  & 2       & 3       & 4       & 5       & 6       & 7       & 8       \\
\midrule
$i=0$            & 0.0400 & -0.1145 & 0.1732  & -0.2082 & 0.2154  & -0.1944 & 0.1492  & -0.0873 \\
1                & 0.0924 & -0.2323 & 0.2573  & -0.1474 & -0.0565 & 0.2694  & -0.3968 & 0.3756  \\
2                & 0.1042 & -0.1979 & 0.0525  & 0.2302  & -0.3988 & 0.2658  & 0.1156  & -0.4673 \\
3                & 0.0897 & -0.1089 & -0.1131 & 0.2743  & -0.0962 & -0.2563 & 0.3350  & 0.0116  \\
4                & 0.0721 & -0.0595 & -0.1299 & 0.1644  & 0.0556  & -0.2012 & 0.0572  & 0.1370  \\
5                & 0.1016 & -0.0416 & -0.2399 & 0.1412  & 0.2804  & -0.2618 & -0.2601 & 0.3821  \\
6                & 0.1016 & 0.0416  & -0.2399 & -0.1412 & 0.2804  & 0.2618  & -0.2601 & -0.3821 \\
7                & 0.0721 & 0.0595  & -0.1299 & -0.1644 & 0.0556  & 0.2012  & 0.0572  & -0.1370 \\
8                & 0.0897 & 0.1089  & -0.1131 & -0.2743 & -0.0962 & 0.2563  & 0.3350  & -0.0116 \\
9                & 0.1042 & 0.1979  & 0.0525  & -0.2302 & -0.3988 & -0.2658 & 0.1156  & 0.4673  \\
10               & 0.0924 & 0.2323  & 0.2573  & 0.1474  & -0.0565 & -0.2694 & -0.3968 & -0.3756 \\
11               & 0.0400 & 0.1145  & 0.1732  & 0.2082  & 0.2154  & 0.1944  & 0.1492  & 0.0873  \\
\botrule
\end{tabular}
\caption{ the first eight kernels for decomposition}
\end{subtable}%

\begin{subtable}[h]{\linewidth}
\centering

\begin{tabular}{ccccccccc}
\toprule
$\psi_{m,i}$     & $m=1$  & 2       & 3       & 4       & 5       & 6       & 7       & 8       \\
\midrule
$i=0$            & 1.0000 & -1.0000 & 1.0000  & -1.0000 & 1.0000  & -1.0000 & 1.0000  & -1.0000 \\
1                & 1.0000 & -0.8182 & 0.5041  & -0.1420 & -0.1748 & 0.3710  & -0.4109 & 0.3063  \\
2                & 1.0000 & -0.6364 & 0.1074  & 0.3103  & -0.4261 & 0.2399  & 0.0752  & -0.2945 \\
3                & 1.0000 & -0.4545 & -0.1901 & 0.4470  & -0.2130 & -0.1833 & 0.3303  & -0.1217 \\
4                & 1.0000 & -0.2727 & -0.3884 & 0.3584  & 0.1203  & -0.3457 & 0.0726  & 0.2596  \\
5                & 1.0000 & -0.0909 & -0.4876 & 0.1345  & 0.3443  & -0.1639 & -0.2596 & 0.1843  \\
6                & 1.0000 & 0.0909  & -0.4876 & -0.1345 & 0.3443  & 0.1639  & -0.2596 & -0.1843 \\
7                & 1.0000 & 0.2727  & -0.3884 & -0.3584 & 0.1203  & 0.3457  & 0.0726  & -0.2596 \\
8                & 1.0000 & 0.4545  & -0.1901 & -0.4470 & -0.2130 & 0.1833  & 0.3303  & 0.1217  \\
9                & 1.0000 & 0.6364  & 0.1074  & -0.3103 & -0.4261 & -0.2399 & 0.0752  & 0.2945  \\
10               & 1.0000 & 0.8182  & 0.5041  & 0.1420  & -0.1748 & -0.3710 & -0.4109 & -0.3063 \\
11               & 1.0000 & 1.0000  & 1.0000  & 1.0000  & 1.0000  & 1.0000  & 1.0000  & 1.0000  \\
\botrule
\end{tabular}
\caption{ the first eight kernels for reconstruction}
\end{subtable}
\caption{Kernel weights for Legendre transformation ($N=12$)}
\label{tab:s2}
\end{table}

\begin{table}[h]
\begin{subtable}[h]{\linewidth}
\centering

\begin{tabular}{ccccccccc}

\toprule

$\varphi_{m,i}$ & $m=1$  & 2       & 3       & 4       & 5       & 6       & 7       & 8       \\
\midrule
$i=0$            & 0.0307 & -0.0882 & 0.1347  & -0.1647 & 0.1754  & -0.1669 & 0.1421  & -0.1062 \\
1                & 0.0577 & -0.1525 & 0.1924  & -0.1623 & 0.0714  & 0.0504  & -0.1630 & 0.2299  \\
2                & 0.0601 & -0.1371 & 0.1116  & 0.0108  & -0.1612 & 0.2478  & -0.2119 & 0.0643  \\
3                & 0.0582 & -0.1115 & 0.0345  & 0.1163  & -0.2070 & 0.1443  & 0.0375  & -0.1995 \\
4                & 0.0518 & -0.0842 & -0.0134 & 0.1433  & -0.1548 & 0.0118  & 0.1587  & -0.1924 \\
5                & 0.0654 & -0.0852 & -0.0709 & 0.2044  & -0.1041 & -0.1575 & 0.2735  & -0.0705 \\
6                & 0.0662 & -0.0529 & -0.1303 & 0.1632  & 0.0780  & -0.2509 & 0.0509  & 0.2642  \\
7                & 0.0440 & -0.0227 & -0.0976 & 0.0723  & 0.0972  & -0.1194 & -0.0641 & 0.1459  \\
8                & 0.0660 & -0.0178 & -0.1610 & 0.0613  & 0.2049  & -0.1176 & -0.2238 & 0.1805  \\
9                & 0.0660 & 0.0178  & -0.1610 & -0.0613 & 0.2049  & 0.1176  & -0.2238 & -0.1805 \\
10               & 0.0440 & 0.0227  & -0.0976 & -0.0723 & 0.0972  & 0.1194  & -0.0641 & -0.1459 \\
11               & 0.0662 & 0.0529  & -0.1303 & -0.1632 & 0.0780  & 0.2509  & 0.0509  & -0.2642 \\
12               & 0.0654 & 0.0852  & -0.0709 & -0.2044 & -0.1041 & 0.1575  & 0.2735  & 0.0705  \\
13               & 0.0518 & 0.0842  & -0.0134 & -0.1433 & -0.1548 & -0.0118 & 0.1587  & 0.1924  \\
14               & 0.0582 & 0.1115  & 0.0345  & -0.1163 & -0.2070 & -0.1443 & 0.0375  & 0.1995  \\
15               & 0.0601 & 0.1371  & 0.1116  & -0.0108 & -0.1612 & -0.2478 & -0.2119 & -0.0643 \\
16               & 0.0577 & 0.1525  & 0.1924  & 0.1623  & 0.0714  & -0.0504 & -0.1630 & -0.2299 \\
17               & 0.0307 & 0.0882  & 0.1347  & 0.1647  & 0.1754  & 0.1669  & 0.1421  & 0.1062  \\
\botrule
\end{tabular}
\caption{ the first eight kernels for decomposition}
\end{subtable}%

\begin{subtable}[h]{\linewidth}
\centering

\begin{tabular}{ccccccccc}
\toprule
$\psi_{m,i}$ & $m=1$  & 2       & 3       & 4       & 5       & 6       & 7       & 8       \\
\midrule
$i=0$         & 1.0000 & -1.0000 & 1.0000  & -1.0000 & 1.0000  & -1.0000 & 1.0000  & -1.0000 \\
1             & 1.0000 & -0.8824 & 0.6678  & -0.3939 & 0.1073  & 0.1447  & -0.3234 & 0.4060  \\
2             & 1.0000 & -0.7647 & 0.3772  & 0.0291  & -0.3218 & 0.4197  & -0.3202 & 0.0950  \\
3             & 1.0000 & -0.6471 & 0.1280  & 0.2933  & -0.4281 & 0.2640  & 0.0436  & -0.2787 \\
4             & 1.0000 & -0.5294 & -0.0796 & 0.4232  & -0.3324 & -0.0218 & 0.2981  & -0.2744 \\
5             & 1.0000 & -0.4118 & -0.2457 & 0.4431  & -0.1350 & -0.2544 & 0.3046  & -0.0149 \\
6             & 1.0000 & -0.2941 & -0.3702 & 0.3776  & 0.0833  & -0.3462 & 0.1172  & 0.2327  \\
7             & 1.0000 & -0.1765 & -0.4533 & 0.2510  & 0.2625  & -0.2841 & -0.1268 & 0.2851  \\
8             & 1.0000 & -0.0588 & -0.4948 & 0.0877  & 0.3621  & -0.1085 & -0.2900 & 0.1247  \\
9             & 1.0000 & 0.0588  & -0.4948 & -0.0877 & 0.3621  & 0.1085  & -0.2900 & -0.1247 \\
10            & 1.0000 & 0.1765  & -0.4533 & -0.2510 & 0.2625  & 0.2841  & -0.1268 & -0.2851 \\
11            & 1.0000 & 0.2941  & -0.3702 & -0.3776 & 0.0833  & 0.3462  & 0.1172  & -0.2327 \\
12            & 1.0000 & 0.4118  & -0.2457 & -0.4431 & -0.1350 & 0.2544  & 0.3046  & 0.0149  \\
13            & 1.0000 & 0.5294  & -0.0796 & -0.4232 & -0.3324 & 0.0218  & 0.2981  & 0.2744  \\
14            & 1.0000 & 0.6471  & 0.1280  & -0.2933 & -0.4281 & -0.2640 & 0.0436  & 0.2787  \\
15            & 1.0000 & 0.7647  & 0.3772  & -0.0291 & -0.3218 & -0.4197 & -0.3202 & -0.0950 \\
16            & 1.0000 & 0.8824  & 0.6678  & 0.3939  & 0.1073  & -0.1447 & -0.3234 & -0.4060 \\
17            & 1.0000 & 1.0000  & 1.0000  & 1.0000  & 1.0000  & 1.0000  & 1.0000  & 1.0000  \\
\botrule
\end{tabular}
\caption{ the first eight kernels for reconstruction}
\end{subtable}
\caption{Kernel weights for Legendre transformation ($N=18$)}
\label{tab:s3}
\end{table}

\begin{table}[h]
\begin{subtable}[h]{\linewidth}
\centering
\begin{tabular}{ccccccccc}
\toprule
$\varphi_{m,i}$ & $m=1$  & 2       & 3       & 4       & 5       & 6       & 7       & 8       \\
\midrule
$i=0$            & 0.0210 & -0.0612 & 0.0966  & -0.1245 & 0.1427  & -0.1502 & 0.1469  & -0.1334 \\
1                & 0.0454 & -0.1241 & 0.1697  & -0.1696 & 0.1234  & -0.0419 & -0.0551 & 0.1446  \\
2                & 0.0430 & -0.1061 & 0.1114  & -0.0508 & -0.0525 & 0.1545  & -0.2093 & 0.1895  \\
3                & 0.0416 & -0.0924 & 0.0674  & 0.0257  & -0.1311 & 0.1796  & -0.1320 & 0.0053  \\
4                & 0.0394 & -0.0792 & 0.0340  & 0.0699  & -0.1511 & 0.1346  & -0.0138 & -0.1370 \\
5                & 0.0544 & -0.0925 & -0.0051 & 0.1507  & -0.1850 & 0.0411  & 0.1723  & -0.2571 \\
6                & 0.0412 & -0.0558 & -0.0401 & 0.1290  & -0.0769 & -0.0858 & 0.1764  & -0.0703 \\
7                & 0.0359 & -0.0421 & -0.0474 & 0.1070  & -0.0311 & -0.1027 & 0.1208  & 0.0142  \\
8                & 0.0458 & -0.0451 & -0.0775 & 0.1296  & 0.0090  & -0.1694 & 0.1097  & 0.1208  \\
9                & 0.0519 & -0.0311 & -0.1143 & 0.1015  & 0.1089  & -0.1753 & -0.0553 & 0.2285  \\
10               & 0.0315 & -0.0117 & -0.0742 & 0.0391  & 0.0864  & -0.0702 & -0.0806 & 0.0977  \\
11               & 0.0488 & -0.0098 & -0.1204 & 0.0340  & 0.1574  & -0.0659 & -0.1799 & 0.1028  \\
12               & 0.0488 & 0.0098  & -0.1204 & -0.0340 & 0.1574  & 0.0659  & -0.1799 & -0.1028 \\
13               & 0.0315 & 0.0117  & -0.0742 & -0.0391 & 0.0864  & 0.0702  & -0.0806 & -0.0977 \\
14               & 0.0519 & 0.0311  & -0.1143 & -0.1015 & 0.1089  & 0.1753  & -0.0553 & -0.2285 \\
15               & 0.0458 & 0.0451  & -0.0775 & -0.1296 & 0.0090  & 0.1694  & 0.1097  & -0.1208 \\
16               & 0.0359 & 0.0421  & -0.0474 & -0.1070 & -0.0311 & 0.1027  & 0.1208  & -0.0142 \\
17               & 0.0412 & 0.0558  & -0.0401 & -0.1290 & -0.0769 & 0.0858  & 0.1764  & 0.0703  \\
18               & 0.0544 & 0.0925  & -0.0051 & -0.1507 & -0.1850 & -0.0411 & 0.1723  & 0.2571  \\
19               & 0.0394 & 0.0792  & 0.0340  & -0.0699 & -0.1511 & -0.1346 & -0.0138 & 0.1370  \\
20               & 0.0416 & 0.0924  & 0.0674  & -0.0257 & -0.1311 & -0.1796 & -0.1320 & -0.0053 \\
21               & 0.0430 & 0.1061  & 0.1114  & 0.0508  & -0.0525 & -0.1545 & -0.2093 & -0.1895 \\
22               & 0.0454 & 0.1241  & 0.1697  & 0.1696  & 0.1234  & 0.0419  & -0.0551 & -0.1446 \\
23               & 0.0210 & 0.0612  & 0.0966  & 0.1245  & 0.1427  & 0.1502  & 0.1469  & 0.1334  \\
\botrule
\end{tabular}
\caption{ the first eight kernels for decomposition}
\end{subtable}%

\begin{subtable}[h]{\linewidth}
\centering
\begin{tabular}{ccccccccc}
\toprule
$\psi_{m,i}$ & $m=1$  & 2       & 3       & 4       & 5       & 6       & 7       & 8       \\
\midrule
$i=0$         & 1.0000 & -1.0000 & 1.0000  & -1.0000 & 1.0000  & -1.0000 & 1.0000  & -1.0000 \\
1             & 1.0000 & -0.9130 & 0.7505  & -0.5333 & 0.2893  & -0.0488 & -0.1594 & 0.3121  \\
2             & 1.0000 & -0.8261 & 0.5236  & -0.1702 & -0.1467 & 0.3542  & -0.4143 & 0.3319  \\
3             & 1.0000 & -0.7391 & 0.3195  & 0.0992  & -0.3679 & 0.4101  & -0.2492 & -0.0095 \\
4             & 1.0000 & -0.6522 & 0.1380  & 0.2848  & -0.4285 & 0.2752  & 0.0280  & -0.2699 \\
5             & 1.0000 & -0.5652 & -0.0208 & 0.3964  & -0.3765 & 0.0659  & 0.2454  & -0.3141 \\
6             & 1.0000 & -0.4783 & -0.1569 & 0.4439  & -0.2539 & -0.1366 & 0.3313  & -0.1772 \\
7             & 1.0000 & -0.3913 & -0.2703 & 0.4372  & -0.0966 & -0.2817 & 0.2826  & 0.0361  \\
8             & 1.0000 & -0.3043 & -0.3611 & 0.3860  & 0.0652  & -0.3445 & 0.1379  & 0.2174  \\
9             & 1.0000 & -0.2174 & -0.4291 & 0.3004  & 0.2076  & -0.3215 & -0.0448 & 0.2937  \\
10            & 1.0000 & -0.1304 & -0.4745 & 0.1901  & 0.3125  & -0.2254 & -0.2065 & 0.2433  \\
11            & 1.0000 & -0.0435 & -0.4972 & 0.0650  & 0.3679  & -0.0808 & -0.3002 & 0.0935  \\
12            & 1.0000 & 0.0435  & -0.4972 & -0.0650 & 0.3679  & 0.0808  & -0.3002 & -0.0935 \\
13            & 1.0000 & 0.1304  & -0.4745 & -0.1901 & 0.3125  & 0.2254  & -0.2065 & -0.2433 \\
14            & 1.0000 & 0.2174  & -0.4291 & -0.3004 & 0.2076  & 0.3215  & -0.0448 & -0.2937 \\
15            & 1.0000 & 0.3043  & -0.3611 & -0.3860 & 0.0652  & 0.3445  & 0.1379  & -0.2174 \\
16            & 1.0000 & 0.3913  & -0.2703 & -0.4372 & -0.0966 & 0.2817  & 0.2826  & -0.0361 \\
17            & 1.0000 & 0.4783  & -0.1569 & -0.4439 & -0.2539 & 0.1366  & 0.3313  & 0.1772  \\
18            & 1.0000 & 0.5652  & -0.0208 & -0.3964 & -0.3765 & -0.0659 & 0.2454  & 0.3141  \\
19            & 1.0000 & 0.6522  & 0.1380  & -0.2848 & -0.4285 & -0.2752 & 0.0280  & 0.2699  \\
20            & 1.0000 & 0.7391  & 0.3195  & -0.0992 & -0.3679 & -0.4101 & -0.2492 & 0.0095  \\
21            & 1.0000 & 0.8261  & 0.5236  & 0.1702  & -0.1467 & -0.3542 & -0.4143 & -0.3319 \\
22            & 1.0000 & 0.9130  & 0.7505  & 0.5333  & 0.2893  & 0.0488  & -0.1594 & -0.3121 \\
23            & 1.0000 & 1.0000  & 1.0000  & 1.0000  & 1.0000  & 1.0000  & 1.0000  & 1.0000  \\
\botrule
\end{tabular}
\caption{ the first eight kernels for reconstruction}
\end{subtable}
\caption{Kernel weights for Legendre transformation ($N=24$)}
\label{tab:s4}
\end{table}




\end{appendices}


\end{document}